\documentclass{clv3}

\usepackage{hyperref}
\usepackage[dvipsnames]{xcolor}
\definecolor{darkblue}{rgb}{0, 0, 0.5}
\hypersetup{colorlinks=true,citecolor=darkblue, linkcolor=darkblue, urlcolor=darkblue}

\issue{0}{0}{2019}

\runningtitle{Detecting Local Insights from Global Labels}

\runningauthor{Allen Schmaltz}

\usepackage{subcaption}
\usepackage{basecommon} 

\usepackage{multirow}
\usepackage{basecommon} 
\usepackage{array}
\newcolumntype{P}[1]{>{\centering\arraybackslash}p{#1}}
\newcolumntype{T}[1]{>{\raggedright\arraybackslash}p{#1}}

\newcommand{\lightgraybox}[1]{{\color{black}\colorbox{lightgray}{#1}} }  % used in some tables
\begin{document}

\title{Detecting Local Insights from Global Labels: Supervised \& Zero-Shot Sequence Labeling via a Convolutional Decomposition}

\author{Allen Schmaltz}
\affil{Department of Epidemiology\\Harvard University\\\texttt{aschmaltz@hsph.harvard.edu}}

\maketitle

\begin{abstract} 
We propose a new, more actionable view of neural network interpretability and data analysis by leveraging the remarkable matching effectiveness of representations derived from deep networks, guided by an approach for class-conditional feature detection. The decomposition of the filter-ngram interactions of a convolutional neural network and a linear layer over a pre-trained deep network yields a strong binary sequence labeler, with flexibility in producing predictions at---and defining loss functions for---varying label granularities, from the fully-supervised sequence labeling setting to the challenging zero-shot sequence labeling setting, in which we seek token-level predictions but only have document-level labels for training. From this sequence-labeling layer we derive dense representations of the input that can then be matched to instances from training, or a support set with known labels. Such introspection with inference-time decision rules provides a means, in some settings, of making local updates to the model by altering the labels or instances in the support set without re-training the full model. Finally, we construct a particular K-nearest neighbors (K-NN) model from matched exemplar representations that approximates the original model's predictions and is at least as effective a predictor with respect to the ground-truth labels. This additionally yields interpretable heuristics at the token level for determining when predictions are less likely to be reliable, and for screening input dissimilar to the support set. In effect, we show that we can transform the deep network into a simple weighting over exemplars and associated labels, yielding an introspectable---and modestly updatable---version of the original model. %, at varying label resolutions. 
\end{abstract}

%%%%%%%%%%%%%%%%%%%%%%%%%%%%%%%%%%%%%%%%%%
%%%% SECTION
%%%%%%%%%%%%%%%%%%%%%%%%%%%%%%%%%%%%%%%%%%

\section{Introduction}
The promise and peril of deep learning in computational linguistics, and AI in general, would seem, on the surface, to be that the strong effectiveness of the large neural networks is unavoidably accompanied by inscrutable model predictions. The models are often right, but when they are wrong, it is difficult to ascertain why, and furthermore, it is typically not obvious how to course-correct a model when errors are discovered, beyond altogether abandoning the model. The non-identifiable \citep[cf.,][]{HwangAndDing-1997-PredictionIntervals,JainAndWallace-2019-AttentionIsNotExplanation} and extraordinarily large number of parameters suggest a lost cause, in general, for tracing model predictions back to particular parameter values, and it would seem then that deep networks are of limited use in settings where interpretability is paramount. However, interestingly, and surprisingly, we show that there is nonetheless a sense in which the deep networks can be leveraged to create a notion of actionable interpretability against the data that is not necessarily possible with simpler, less expressive models alone, and may yield precisely the characteristics desired in certain real-world applications. By leveraging the strong pattern matching behavior and the dense representations of the deep networks, we can form a mapping between test instances and training instances with known labels, which enables introspection of the model with respect to the data. In some settings, we can then update the model by updating the data and labels in these mappings. Interestingly, in this way, the application of deep neural networks begins to resemble some of the classic instance-based and metric learning methods from machine learning, as well as the exemplar systems \citep{Clark-1990-Early-Inelligent-Systems} from an earlier era of AI, but with less dependence on human-mediated feature engineering, which may prove critical for applications with high-dimensional input, at the very least as tools for data analysis. 

A model for analyzing a natural language dataset ideally needs some facility for class-conditional feature detection at the word level. However, the compositional, high-dimensional nature of language makes feature detection a challenging endeavor, with further empirical complications arising from the need to label at a granularity that is typically more fine-grained than many existing human-annotated datasets. We propose and demonstrate that a single-layer, one-dimensional, kernel-width-one max-pooled convolutional neural network and a linear layer, as the final layer of a network, can be trained for document-level classification, and then decomposed in a straightforward way to produce token-level labels. This particular set of operations over a CNN and a linear layer yields flexibility in learning and predicting at disparate label resolutions and is efficient and simple to calculate and train. It can readily replace the standard final linear layer often used for classification in Transformer models \cite{VaswaniEtAl-2017-Transformer}, adding the properties described here. We empirically show across tasks, using datasets that have token-level labels for verification, that it yields surprisingly sharp token-level binary detections even when trained at the document level, when the input to the layer is a large, masked-language-model-trained \textsc{BERT} model \citep{DevlinEtAl-2018-BERT}.  

We find feature detection in this way to be a useful tool for analyzing datasets, detecting rather subtle distributional differences within documents that can be otherwise challenging to find at scale. Further, we show that the CNN filter applications corresponding to the token-level predictions are effective dense representations of the model predictions, with which we can form a mapping between test predictions and instances with known labels. We find qualitatively and quantitatively that the matches correspond to similar features in similar contexts, at least when the distances between representations are low. Finally, without loss of predictive effectiveness, we can altogether replace the model's output with a simple weighting over exemplar representations, converting the deep network into a K-nearest neighbor (K-NN) model, with concomitant benefits for interpretability, and straightforward heuristics for detecting domain-shifted and out-of-domain data. 

In summary, this work contributes the following new approaches:
\begin{enumerate}
\item We present a new, effective model for supervised and zero-shot binary sequence labeling. We evaluate on token-level annotations for grammatical error detection and diff annotations on a sentiment dataset, detecting both sentiment features and surprisingly, also subtle re-annotation artifacts. 
\item We propose a method for data and model analysis via dense representation matching, \textit{exemplar auditing}, enabled by our binary sequence labeling method, creating inference-time decision rules linking feature-level exemplar representations and associated predictions from test with representations from a support set with known labels. We show that in some settings we can make local updates to the model by updating the data and labels in the support set without re-training the full model.  
\item We approximate the model's token-level output with a K-NN over the support set that is at least as effective as the original model, and can be used as an interpretable substitute for the original model. Incorrect model predictions tend to also be more difficult to approximate; our proposed approach yields simple, understandable heuristics at the token level for determining when predictions are less likely to be reliable, and for screening input unlike that seen in the support set. 
\end{enumerate}

We proceed by first introducing the notation for the tasks across label resolutions (Section~\ref{sec:task}) and the core methods (Section~\ref{sec:methods}) used across all experiments, and then we apply these ideas to three tasks. First, we demonstrate effectiveness on the challenging, well-defined error detection task (Section~\ref{sec:error-detection}), which enables careful examination of the behavior using available token-level labels. Next, we use sentiment data that has been usefully re-annotated via local changes (Section~\ref{sec:sentiment}) to further examine updating the support set over domain-shifted data, and to motivate and analyze our approach for constraining out-of-domain data in the context of an existing approach for robust classification. Finally, we also use these sentiment datasets to examine the model's ability to detect subtle distributional changes across re-annotated and original data (Section~\ref{sec:local-annotation-detection}), discovering features that are not readily detectable at scale without model-based assistance.  

%%%%%%%%%%%%%%%%%%%%%%%%%%%%%%%%%%%%%%%%%%
%%%% SECTION
%%%%%%%%%%%%%%%%%%%%%%%%%%%%%%%%%%%%%%%%%%
\section{Tasks}\label{sec:task}

Given a document, which may consist of a single sentence, we seek binary labels over the words in the document. For learning such a model, we may be given training examples with associated labels for each of the ``words''\footnote{Hereafter, we will tend to use ``token'' instead of ``word'', as the lowest resolution of the input will be determined by the tokenization scheme of the particular dataset.}, which is the standard fully-supervised binary sequence labeling setting, or we may only be given document-level labels, which is the zero-shot binary sequence labeling setting. This latter setting corresponds to notions of feature detection for document-level classification models, enabling quantitative evaluation when given token-level labeled held-out data.

%\rvx and \vx look similar here, so we just stick to \vx
\paragraph{Supervised Binary Sequence Labeling} Specifically, in the standard fully-supervised sequence labeling setting, we are given a training dataset $\sD^*=\{(\vx_{d}, \vy_{d}) \given 1\le d \le |\sD^*|\}$ of $|\sD^*|$ documents paired with their corresponding token-level ground-truth labels. Each of $N$ tokens in a document, $\vx=\evx_1,\ldots,\evx_n,\ldots,\evx_N$, has a known token-level label, $\evy_n\in\{-1,1\}$. We seek a learned mapping, $\vx \mapsto \hat{\vy}$, for predicting the labels for a given document: At inference, we are given a new, previously unseen document instance, $\vx_{|\sD^*|+1}$, over which we predict $\hat \vy_{|\sD^*|+1}=\hat{\evy}_1,\ldots,\hat{\evy}_n,\ldots,\hat{\evy}_N$, the token-level labels for each token in the document. We will subsequently drop the subscript label, ``${|\sD^*|+1}$'', on test-time instances when the distinction from training is otherwise unambiguous. We aim to minimize the distance between the predicted $\hat \vy$ and the ground-truth $\vy$. 

Throughout we use $^*$ to indicate a dataset includes, or a model otherwise has access to, token-level labels. Otherwise, the label signal is limited to the document level, with the exception of clearly indicated reference experiments simply tuning the decision boundary of document-level models with a limited number of token-level labels. 

\paragraph{Document-level Binary Classification} The zero-shot binary sequence labeling setting also seeks to predict token-level labels, but in this case, the model is only trained with \textit{document-level} labels, as in standard document-level classification, the notation for which we introduce here. We are given a training dataset $\sD=\{(\vx_d, Y_d) \given 1\le d \le |\sD|\}$ of $|\sD|$ documents paired with their corresponding document-level ground-truth labels. Token-level labels are not present in $\sD$. At inference, we seek to predict $\hat Y$ given a new, unseen document $\vx$, via the learned mapping $F: \vx \mapsto \hat Y$. We aim for $\hat Y$ to be close to the true document-level label, $Y\in \{-1,1\}$.%\footnote{Note the distinction: $Y$ is a document-level label, whereas $y_n$ is a token-level label.}

\paragraph{Zero-shot Binary Sequence Labeling} The zero-shot binary sequence labeling models have access to the same training dataset $\sD$ as in the standard document-level classification task. However, at inference, we then seek to predict the \textit{token-level} labels, $\hat \vy$, for each token in the new document instance $\vx$, via a mapping $\vx \mapsto \hat \vy$, even though we can only query the document-level labels of $\sD$ during training. In other words, the learning signal is the same for document-level classification and zero-shot sequence labeling, but the inference-time task is the same in the zero-shot sequence labeling and fully-supervised sequence labeling settings. \\

We will be primarily concerned with analyzing the sequence labeling settings. We also report document-level classification results for a subset of the zero-shot sequence labeling models, illustrating how the proposed token-level predictions can be used to analyze and constrain typical text datasets that only have labels at the document level, rather than at finer-grained resolutions, at least at scale. 

%%%%%%%%%%%%%%%%%%%%%%%%%%%%%%%%%%%%%%%%%%
%%%% SECTION
%%%%%%%%%%%%%%%%%%%%%%%%%%%%%%%%%%%%%%%%%%
\section{Methods}\label{sec:methods}

We propose a new method for class-conditional feature detection from a large, expressive deep network that enables the interlinked view of interpretability, constrained inference, and updatability via an external database introduced in this work. We demonstrate that a particular max-pool attention-style mechanism from a CNN and a linear layer over a deep network enables the following:  

\begin{enumerate}
\item We show that we can derive token-level predictions across the full document, $f(\evx_1),\ldots,f(\evx_n),\ldots,f(\evx_N)$, from the document-level prediction, $F(\vx)$. This decomposition provides flexibility in learning and analyzing at varying label resolutions.  
\item We further show that the token-level predictions can themselves be approximately decomposed via $f(\evx_n)\approx f(\evx_n)^{KNN}$, where $f(\evx_n)^{KNN}$ is an explicit weighting over a set of nearest exemplar representations and their associated labels and predictions.
\end{enumerate}

We proceed by first introducing the base document-level classifier (Section~\ref{sec:methods-doc-level-classifier}). We then introduce the approach for deriving token-level predictions from the document-level classifier (Section~\ref{sec:methods-cnn-linear-decomposition}) and how this can be used for supervised labeling (Section~\ref{sec:methods-supervised-labeling}); yields flexibility in adding task-specific priors (Section~\ref{sec:methods-min-max-constraints}); and provides a means of aggregate feature extraction for analyzing datasets (Section~\ref{sec:methods-feature-extraction}). Next, we introduce the approach for mapping a test-time prediction to a database of exemplars by leveraging dense representations coupled to the class-conditional feature detection (Section~\ref{sec:methods-exa}), before introducing the K-NN approximations (Section~\ref{sec:methods-k-nn}).\footnote{Our replication code is publicly available at \url{https://github.com/allenschmaltz/exa}.}

%The model, trained via standard Empirical Risk Minimization approaches (e.g., via a cross-entropy loss without instance-level re-weighting) can be viewed as a function (here, a deep parametric network) 

\subsection{CNN Binary Classifier Over a Deep Network: Document-Level Predictions}\label{sec:methods-doc-level-classifier}

We use a CNN architecture similar to that of \citet{Kim-2014-CNN} over a pre-trained Transformer model \citep{DevlinEtAl-2018-BERT} and fine-tuned word embeddings as our document-level classifier, $F$. Each token $\evx_n \in \vx$ in the document, including padding symbols as necessary, is represented by a $D$-dimensional vector, $\vt_n=(\ve^{BERT}, \ve^{word})$, the concatenation of the top hidden layer(s) of a Transformer and a vector of word embeddings, $D=\left| \ve^{BERT} \right | + \left| \ve^{word} \right |$. The convolutional layer is then applied to this $\reals^{D \times N}$ matrix, using a filter of width $Q$, sliding across the dense vectors corresponding to the $Q$-sized n-grams of the input. The convolution results in a feature map $\vh_m \in \reals^{N-Q+1}$ for each of $M$ total filters.% Note that each of the filters has a bias and $(D \cdot K)$ weights. 

We then compute
\begin{align*}
g_m &= \max \relu(\vh_m), 
\end{align*}
a $\relu$ non-linearity followed by a max-pool over the n-gram dimension resulting in $\vg \in \reals^M$. A final linear fully-connected layer, $\mW \in \reals^{C \times M}$, with a bias, $\vb \in \reals^C$, followed by a softmax, produces the output distribution over $C$ class labels, $\vo \in \reals^C$: 
\begin{align*}
\boldo &= \text{softmax} (\boldW\boldg + \boldb). 
\end{align*}

The base model is trained for document classification with a standard cross-entropy loss. We primarily use a filter width of 1, $Q=1$. In experiments with multiple filter widths, we concatenate the output of the max-pooling prior to the fully-connected layer.

\begin{figure}[!htbp]
    \centering
    \includegraphics[scale=0.45]{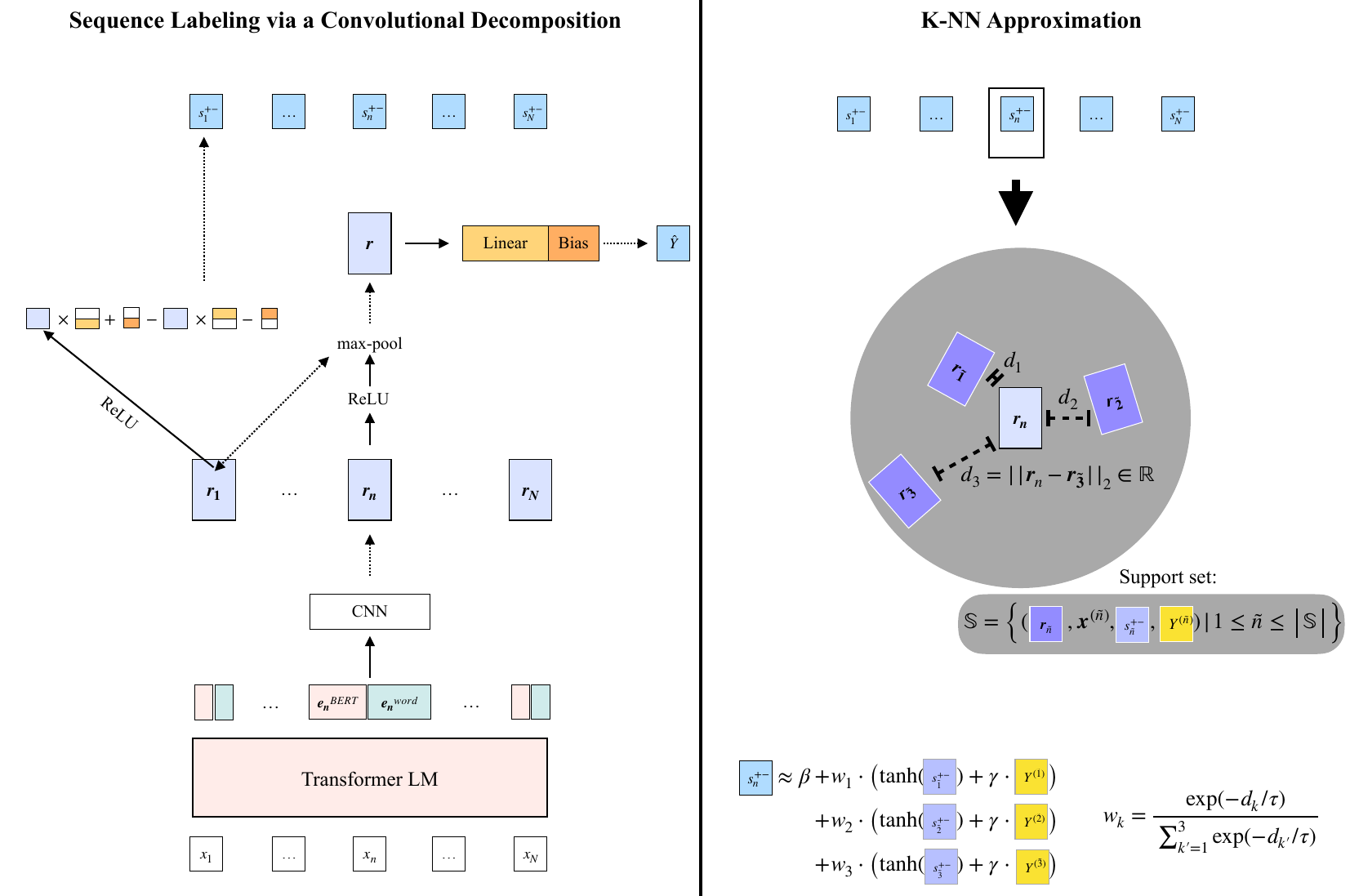}
    \caption{High-level overview of the proposed methods. We derive token-level predictions from a model trained with document-level labels via the decomposition of a max-pooled, kernel-width-one CNN and a linear layer over a large Transformer language model (\textit{left}). These token-level predictions can themselves be approximated as an interpretable weighting over a support set with known labels (\textit{right}, where $K=3$ in the illustration) by leveraging the CNN's feature-specific, summarized representations of the deep network to measure distances to the support set.}
    \label{fig:overview}
\end{figure}

\subsection{Zero-shot Sequence Labeling with a CNN Binary Classifier: From Document-Level Labels to Token-Level Labels}\label{sec:methods-cnn-linear-decomposition} 

The matrix multiplication of the output of the max-pooling operation with the fully-connected layer can be viewed as a weighted sum of the most relevant filter-ngram interactions for each prediction class. This can be deterministically decomposed to produce predictions at the resolution of the CNN's input for each class. %More specifically, each term in $\mW\vg$ contributing to the negative class prediction (for the purposes here, the class at index 1), $W_{1,1}\cdot g_1,\ldots,W_{1,M}\cdot g_M$, can be deterministically traced back to its corresponding filter and the tokens in the window on which the filter was applied. For each token in the input, we assign a \textit{negative class contribution} score by summing all applicable terms in $\mW_{1,1:M}\vg$ in which the token interacted with a filter that survived the max-pooling operation. Some tokens may have a resulting score of zero. Similarly, we assign a \textit{positive class contribution} score by summing all applicable terms in $\mW_{2,1:M}\vg$. We now have a decision boundary for each token, assigning the positive class when the \textit{positive class contribution} score is greater than the \textit{negative class contribution} score. To account for the bias in the fully-connected layer, we add the bias in the fully-connected layer contributing to the negative class prediction to the \textit{negative class contribution} scores, and we add the bias in the fully-connected layer contributing to the positive class prediction to the \textit{positive class contribution} scores.%\footnote{The aggregated bias, a constant across sentences, amounts to an offset on the token-level decision boundary. Aside from the size of the CNN itself, this is perhaps the main parameter end-users may want to examine adjusting on held-out data, if available, for their particular dataset if not using a task-specific loss. Additionally, if the classifier is weak, it may be helpful to constrain prediction of token labels to sentences that are predicted to be of the target class at the sentence level.}  % consider adding this back, noting that the MM loss serves as a means of adjusting this parameter (perhaps mention below in the MM loss section)
Specifically, we use the notation
\begin{align*}
n_m &= \argmax \relu(\vh_m), 
\end{align*}
to identify the index into the feature map $\vh_m$ that survived the max-pooling operation, which corresponds to the application of filter $m$ starting at index $n_m$ of the input (i.e., the set $\{n_m,\ldots,n_m+(Q-1)\}$ contains all of the indices of the input covered by this particular application of the filter of width $Q$). We then have a corresponding negative contribution score $s^{-}_n \in \reals$ for each input token:
\begin{align*}
s^{-}_n &= \left(\sum_{m=1}^{M}W_{1,m}\cdot g_m\cdot\sum_{q=1}^Q [n=n_m+(q-1)] \right) + b_1, 
\end{align*}
where we have used an Iverson bracket for the indicator function. The corresponding positive contribution score $s^{+}_n$ is analogous:
\begin{align*}
s^{+}_n &= \left(\sum_{m=1}^{M}W_{2,m}\cdot g_m\cdot\sum_{q=1}^Q [n=n_m+(q-1)] \right) + b_2. 
\end{align*}

This decomposition then affords considerable flexibility in defining loss constraints to bias the filter weights according to the granularity of the available labels, and/or according to other priors we may have regarding our data.

\subsection{Supervised Sequence Labeling}\label{sec:methods-supervised-labeling} 

We can use the aforementioned decomposition to fine-tune against token-level labels, when available. We subtract the \textit{negative class contribution} scores from the \textit{positive class contribution} scores, passing the result through a sigmoid transformation for each token. We minimize a binary cross-entropy loss, averaged over the non-padding tokens in the mini-batch:
\begin{align*}
\mathcal{L}_n &= -y'_n\cdot \log \sigma(s^{+-}_n) - (1-y'_n)\cdot \log (1-\sigma(s^{+-}_n)), 
\end{align*}
where $s^{+-}_n = s^{+}_n-s^{-}_n$ and $y'_n \in \{0,1\}$ is the corresponding true token label, transformed via: 

\begin{align*}
y'_n = \begin{cases} 
      1 & \text{if} ~y_n=1 \\
      0 & \text{if} ~y_n=-1. \\
   \end{cases}
\end{align*}

For inference, token-level detection labels are determined in the same manner as in the zero-shot setting. %PyTorch autograd efficiently handles backpropagation through the constructed structure. 

\subsection{Task-Specific Zero-Shot Loss Constraints: Min-Max}\label{sec:methods-min-max-constraints}

The base zero-shot formulation is appealing because it only requires labels at the document level, and does not entail additional losses nor other constraints beyond the standard classifier. This mechanism also enables adding task-specific constraints, where applicable, to bias the token contributions based on priors we may have about our data. For example, \citet{ReiAndSogaard-2018-ZeroShotSeq} proposes a min-max squared loss constraint for grammatical error detection. We can capture this idea in our setting in the following manner by fine-tuning the CNN parameters with the following binary cross-entropy losses: %As we demonstrate below, this can be quite effective across tasks, serving as an efficient and useful means of backing out token-level labels from document-level labels. 
\begin{align*}
\mathcal{L}_{min} &= -\log (1-\sigma(s^{+-}_{min})), 
\end{align*}
where $s^{+-}_{min}=\min(s^{+-}_1,\ldots,s^{+-}_n,\ldots,s^{+-}_N)$ is the smallest combined token contribution in the sentence; and
\begin{align*}
\mathcal{L}_{max} &= -Y'\cdot \log \sigma(s^{+-}_{max}) - (1-Y')\cdot \log (1-\sigma(s^{+-}_{max})), 
\end{align*}
where $s^{+-}_{max}=\max(s^{+-}_1,\ldots,s^{+-}_n,\ldots,s^{+-}_N)$ is the largest combined token contribution in the sentence and $Y'$ is the true document-level label, $Y$, transformed to be in $\{0,1\}$. These two losses are then averaged together over the mini-batch. 

The intuition is to encourage correct sentences to have aggregated token contributions less than zero (i.e., no detected errors), and to encourage sentences with errors to have at least one token contribution less than zero and at least one greater than zero (i.e., to encourage even incorrect sentences to have one or more correct tokens, since errors are, in general, relatively rare).  

\subsection{Aggregate, Comparative Feature Extraction}\label{sec:methods-feature-extraction}
From the token-level contributions, we can then score spans of text, from n-grams to full sentences and documents, serving as a type of feature extractor for each class. We can aggregate token contributions across spans of text, which can have the effect of comparative, extractive summarization, an additional useful view of a dataset under a model. Here we assign scores to the negative class n-grams of size $z$ as follows\footnote{We drop the constant bias term since we are ranking negative and positive class n-grams separately.}: 
\begin{align*}
\ngram_{n:n+(z-1)}^{-} &= \sum\limits_{i=n}^{n+(z-1)} (s^{-}_i - b_1). 
\end{align*}
The score for the full document is then $\ngram_{1:N}^{-}$. The negative class n-grams are only calculated from documents for which the document-level model predicts the document as being negative. In our analysis below, we consider unigram to 5-gram scores that are summed, $\totalngram_{n:n+(z-1)}^{-}$, or averaged, $\meanngram_{n:n+(z-1)}^{-}$, over the number of occurrences. Similarly, each document is scored by calculating $\ngram_{1:N}^{-}$, and then optionally, normalizing by the document length. The corresponding scores for the positive class, $\ngram_{n:n+(z-1)}^{+}$, are calculated in an analogous manner. 

With the true document-level labels, we can then identify the n-grams and documents most salient for each class under this metric, and just as importantly for many applications, the n-grams and documents that the model misclassifies.

%This type of feature extraction serves as a type of \textit{comparative} summarization, with the scoring of full sentences resembling extractive summarization. We would suggest that this is often the typical scenario for real-world use-cases for text summarization and text analysis: Typically when large-scale automatic summarization is needed, the de-facto goal is \textit{comparative} summarization, comparing the contents of one group of documents to one or more other distinct groups of documents (e.g., highlighting the salient points from negative and positive reviews or comments for a particular product or policy proposal).

\subsection{Exemplar Auditing: Inference-time Decision Rules \& Data/Model Introspection via Dense Representation Matching}\label{sec:methods-exa}

We can view each token-level prediction, $f(\evx_n) = s^{+-}_n$, as the composition $f=u\circ v$, where $v: \ve_n \in \reals^D \mapsto \vr_n \in \reals^M$ and $u: \vr_n \in \reals^M \mapsto s^{+-}_n \in \reals$. The mapping $v$ takes as input the word embeddings and hidden layers of the deep network corresponding to the particular token and produces a dense representation, a distilled summarization of the expressive deep network at the local level which we refer to as an \textit{exemplar} representation, derived from the CNN filter applications corresponding to the token.\footnote{We use the term ``exemplar'' rather than ``prototype'', as we use these representations directly, unique to each feature, rather than as class-based centroids.}

More specifically, with $Q=1$, for each token we have a vector
\begin{align*}
\vr_n = \evh_{1,n},\ldots,h_{m,n},\ldots,h_{M,n}, 
\end{align*}
consisting of the components from each of the $M$ feature maps corresponding to the token at index $n$. With our model, the mapping $u$ is then the max-pool, 
$\relu$, and the corresponding weights of the final fully-connected layer that produce $s^{+-}_n$.

Over a set of instances with known labels containing $\left| \sS \right|$ tokens, we can then form what we term a \textit{support set}:
\begin{align*}
\sS = \left\{ (\vr_{\tilde{n}}, \vx^{({\tilde{n}})}, s^{+-}_{\tilde{n}}, Y^{({\tilde{n}})}) \given 1\le \tilde{n} \le \left| \sS \right| \right\}, 
\end{align*}

a database of meta-data associated with the model's predictions over the document instances for each token index $\tilde{n}$: The token-level representation $\vr_{\tilde{n}}$, the associated document $\vx^{({\tilde{n}})}$, the prediction $s^{+-}_{\tilde{n}}$, and the ground-truth document-level label $Y^{({\tilde{n}})}$. When token-level labels are available, we additionally add $y_{\tilde{n}}$. We treat each $\tilde{n}$ as uniquely describing a single token in the database. The set of documents in the support set and that of the model's training set can be identical, partially overlapping, or even disjoint. 

To aid in analyzing the decision-making process of the model, as well as to explore the characteristics of the data, we can then relate a new test instance to this support set by matching against representations, searching\footnote{We restrict our experiments to exact search, which is nonetheless reasonably fast using GPUs at this scale, to avoid introducing another source of variation, but approximate search could be used in practice for larger support sets.} for the index $\tilde{n}$ that minimizes the Euclidean distance between $\vr_{\tilde{n}}$ and that of the test token's vector $\vr_{n}$:
\begin{align*}
 \argmin_{\tilde{n}} \| \vr_{n} - \vr_{\tilde{n}} \|_2.
\end{align*}

This connection enables inference-time decision rules with which we can inspect and constrain predictions, which we refer to as \textit{exemplar auditing}. We will use the label \textsc{ExAG} for the rule in which positive token-level predictions are only admitted when the token-level prediction of the corresponding exemplar token from the support set matches that of the test token, and the exemplar's document has a positive ground-truth label: $s^{+-}_n>0 \wedge s^{+-}_{\tilde{n}}>0 \wedge Y^{({\tilde{n}})}=1$. Similarly, we use the label \textsc{ExAT} when token-level ground-truth labels are available in the support set: $s^{+-}_n>0 \wedge s^{+-}_{\tilde{n}}>0 \wedge y_{\tilde{n}}=1$. In this way, updates to the support set can be a means of making local updates to the model without modifying the parameters of the original model, including in some cases for domain-shifted data over which the original model is otherwise a weak predictor, provided the dense representations yield adequate matching effectiveness across the new domain. The distances to the matches can also be used for constraining predictions, which we consider in the context of the K-NN approximations described next.

\subsection{K-Nearest Neighbor Model Over Exemplar Representations}\label{sec:methods-k-nn}

The inference-time decision rules are appealing, as once a dense search infrastructure is in place, they are easy to implement and for end-users and auditors to understand: \textit{If a prediction does not resemble that of its nearest matched exemplar, as via a large distance and/or label and prediction discrepancies, reject the prediction and send the decision to a human for adjudication}. Additionally, because the original model's output is used for non-rejected predictions, the prediction effectiveness is guaranteed to be the same as that of the original model for the non-rejected predictions. However, in some settings where explainability is paramount, we may require the stronger sense of fully describing a prediction as a weighting over exemplars from the support set. Interestingly, we show that we can construct a K-NN from a simple transformation of the predictions and class labels of the nearest K exemplars that closely matches the sign directions of the original prediction and is at least as strong a predictor on the metrics over the ground-truth. 

We consider one primary formulation and two additional variations for further analysis. We aim to keep the number of parameters to a minimum to avoid over-fitting; since our goal is to simply reproduce the sign of the original prediction, rather than to construct a significantly larger or more expressive model; and since we seek a weighting that is easily inspectable by an end-user. %For comparison purposes, we train all models in a similar fashion via gradient descent, as described below; however, a simple grid-search, or using the pseudoinverse as with regression, may be sufficient for the variation with only two parameters. 

We seek a simple function that approximates the original model's prediction for a token $\evx_n$ as a weighting over the support set: 
\begin{align*}
\hat y_n &= \sgn\left( f(\evx_n)\right) = \sgn\left(s^{+-}_n\right) \approx \\ 
\hat y^{KNN}_n &= \sgn\left( f(\evx_n)^{KNN}\right) = 
\sgn\left( \beta +
\sum_{k \in \argKmin\limits_{\tilde{n}} || \vr_{n} - \vr_{\tilde{n}} ||_2} w_k \cdot \left({\rm{tanh}}(s^{+-}_k)+\gamma \cdot Y^{(k)} \right)
\right),
\end{align*}

where $\gamma \in \reals$ and $\beta \in \reals$ are parameters learned via gradient descent; with $K$ treated as a hyper-parameter; and $\sgn$ is the binary threshold function
\begin{align*}
\sgn(x) = \begin{cases} 
      1 & \text{if} ~x > 0 \\
      -1 & \text{if} ~x \leq 0. \\
   \end{cases}
\end{align*}

The three considered variations differ in their particular formulation of $w_k$, detailed below, but in all cases $\sum w_k = 1, w_k \in [0, 1]$. We take $s^{+-}_k$ to mean the token-level prediction of the $k^{th}$ nearest exemplar in the support set, and $Y^{(k)} \in \{-1,1\}$ as the document-level label associated with the \textit{document} to which the $k^{th}$ exemplar belongs in the support set. When token-level labels are available, as with the fully-supervised setting, we replace $Y^{(k)}$ with $y_k \in \{-1,1\}$, the ground-truth token-level label associated with the $k^{th}$ exemplar. The $\gamma \cdot Y^{(k)}$ term is in effect a class-specific bias offset given the matched document, and the $\gamma \cdot y_k$ variation directly balances the signal from the true token-level label and the prediction. The predictions and exemplar matchings are at the \textit{token level}, but importantly $\vr$ is a representation of the token that encodes contextual dependencies over the full input, as a result of the deep network.  %As a shorthand label, we use $f(\cdot)^{KNN}$ to mean the un-thresholded output of the K-NN, and similarly, for the output of the original model, $f(\cdot)=s_n^{+-}$ for some $n$. 

\subsubsection{Distance-weighted K-NN (\textsc{KNN\textsubscript{dist.}})} Our main form for $w_k$ accounts for the relative distribution of distances in the top-$K$:

\begin{align*}
w_k =
      \frac{\exp\left(-|| \vr_{n} - \vr_{k} ||_2/\tau \right )}{
             \sum\limits_{k' \in \argKmin\limits_{\tilde{n}} || \vr_{n} - \vr_{\tilde{n}} ||_2} \exp\left(-|| \vr_{n} - \vr_{k'} ||_2/\tau \right )
             },
\end{align*}

where $\tau \in \reals$ is the single additional learnable parameter. \textit{We separately use the raw, unnormalized distance to the nearest match as an exogenous factor to consider when assessing the reliability of the predictions.}

We train the K-NN's parameters with a binary cross-entropy loss, after having trained the original model, the parameters of which remained fixed, by minimizing the difference between the original model's output and the K-NN's output: 

\begin{align*}
\mathcal{L}^{KNN}_n &= -\sigma(s^{+-}_n)\cdot \log \sigma\left( f(\evx_n)^{KNN}\right) - (1-\sigma(s^{+-}_n))\cdot \log \left(1-\sigma\left( f(\evx_n)^{KNN}\right)\right). 
\end{align*}

$\mathcal{L}^{KNN}_n$ is averaged over mini-batches constructed from the tokens of shuffled documents. Across datasets, we treat the original training set, or a subset thereof, as the support set during training, and we randomly split the held-out dev set into two sets: We use half of the data for learning via $\mathcal{L}^{KNN}$, and the other half serves as the held-out \textsc{KNN dev} set. We choose the epoch that minimizes 
\begin{align*}
\delta^{KNN} = \sum\limits_{n \in \text{dev}} [\sgn\left(s^{+-}_n\right)\ne \sgn\left( f(\evx_n)^{KNN}\right) ],
\end{align*}
the total number of prediction discrepancies between the original model and the K-NN approximation over the \textsc{KNN dev} set. During training, if the immediately preceding epoch does not yield the minimal $\delta^{KNN}$ among all epochs, we subsequently only calculate $\mathcal{L}^{KNN}$ for the tokens with prediction discrepancies until a new minimum $\delta^{KNN}$ is found, or the maximum number of epochs is reached.

\subsubsection{Constraint-weighted K-NN (\textsc{KNN\textsubscript{const.}})} We additionally consider a variation to assess the significance of the relative distances by dropping the dependence of $w_k$ on the distances, at the expense of adding $K$ additional learned parameters: 

\begin{align*}
w_k =
      \frac{\exp\left(\bar{\evw}_k/\tau \right )}{
             \sum^K\limits_{k'=1} \exp\left(\bar{\evw}_{k'}/\tau \right )
             },
\end{align*}

with $\tau \in \reals$ and $\bar{\vw} \in \reals^K$.

To avoid overfitting and to encourage the normalized weights to be of decreasing magnitude, $w_k \ge w_{k+1}$, a prior that the closer exemplars should be more prominent in the prediction as with the distance-weighted version above, we add additional loss constraints when training this version: 

\begin{align*}
\mathcal{L}^{\text{KNN}\textsubscript{const.}}_{mm} = \frac{1}{K+1}\left(-\log (1-\sigma(\bar{w}_{min}))-\log \sigma(\bar{w}_{max}) - \sum^{K-1}\limits_{k=1} \log \sigma(\bar{w}_{k}-\bar{w}_{k+1})\right), 
\end{align*}

where $\bar{w}_{min}=\min(\bar{w}_1,\ldots,\bar{w}_k,\ldots,\bar{w}_K)$ is the smallest element of $\bar{\vw}$, the unnormalized weights; $\bar{w}_{max}=\max(\bar{w}_1,\ldots,\bar{w}_k,\ldots,\bar{w}_K)$ is the largest element of $\bar{\vw}$; and the final term encourages decreasing weights. The unnormalized weights, $\bar{\vw}$, are initialized to be decreasing. The final combined loss in a mini-batch for this model is then

\begin{align*}
\mathcal{L}^{\text{KNN}\textsubscript{const.}} = \frac{1}{2} \left( \mathcal{L}^{\text{KNN}\textsubscript{const.}}_{mm} + \frac{1}{|\text{batch}|} \sum^{|\text{batch}|}\limits_{n \in \text{batch}} \mathcal{L}^{KNN}_n \right).
\end{align*}

\subsubsection{Equally-weighted K-NN (\textsc{KNN\textsubscript{equal}})} Finally, we consider $w_k = \frac{1}{K}$. An advantage of this approach is that it requires learning and interpreting only two parameters,  $\gamma$ and $\beta$; it is just a simple transformation of the nearest exemplar predictions and associated labels. A disadvantage is that even relatively far exemplars will play an equal role in the final K-NN prediction. In this way, an interpretation of the model is obligated to equally consider even the farthest exemplars, which requires an end-user to examine the full set of size $K$, some members of which may have near-zero weights in the above alternatives that explicitly enforce a ranking. For comparison purposes, we train this version via gradient descent with $\mathcal{L}^{KNN}$, as with \textsc{KNN\textsubscript{dist.}} above. %; however, a simple grid-search, or using the pseudoinverse as with regression, may be sufficient, as well.%\footnote{Some care would be needed in showing the output to end-users. An argument could be made that the set of nearest exemplars would need to be shuffled, rather than ranked by distance, to avoid end-users from indirectly only considering that the nearest exemplars define the final prediction.}

%%%%%%%%%%%%%%%%%%%%%%%%%%%%%%%%%%%%%%%%%%
%%%% SECTION
%%%%%%%%%%%%%%%%%%%%%%%%%%%%%%%%%%%%%%%%%%
\section{Grammatical Error Detection}\label{sec:error-detection}

The task of grammatical error detection is to detect the presence or absence of grammatical errors in a sentence\footnote{For the FCE dataset, each ``document'' consists of a single sentence.} at the token level.

\subsection{Grammatical Error Detection: Experiments}

We evaluate detection in both the zero-shot and fully-supervised sequence labeling settings, comparing the behavior of the proposed sequence labeling layer to previous models, as well as investigating the behavior of the inference-time decision rules and the K-NN approximations.

\subsubsection{Data: FCE} 

We follow past work on error detection and use the standard training, dev, and test splits of the publicly released subset of the First Certificate in English (FCE) dataset \cite{YannakoudakisEtal-2011-FCEdataset,ReiAndYannakoudakis-2016-NeuralBaselines}\footnote{\url{https://ilexir.co.uk/datasets/index.html}}, consisting of 28.7k, 2.2k, and 2.7k labeled sentences, respectively.
%28,731, 2,222, 2,720 labeled sentences, respectively.

\subsubsection{Data: Domain-shifted News Data} In a real deployment, we might reasonably expect an error detection model to encounter well-formed, correct documents from another domain, over which we would want the model to be robust to false positives. To emulate this scenario, we also consider a series of experiments in which we augment the FCE dataset with sentences from the news-oriented One Billion Word Benchmark dataset \cite{ChelbaEtAl-2013-OneBillionWordBenchmark}, which are assigned negative class ($Y=-1$) sentence-level labels. We augment the FCE training set with a sample of 50,000 sentences (\textsc{FCE+news50k}) and add a disjoint sample of 2,000 sentences to the FCE test set for evaluation (\textsc{FCE+news2k}).

\subsubsection{Models} 

\paragraph{\textsc{uniCNN+BERT} Model} Our primary model uses a filter width of 1 with 1000 filter maps, $Q=1, M=1000$. The CNN layer takes as input, for each token, the top four hidden layers of the large, pre-trained Bidirectional Encoder Representations from Transformers (BERT\textsubscript{LARGE}) model of \citet{DevlinEtAl-2018-BERT}, a multi-layer bidirectional Transformer \cite{VaswaniEtAl-2017-Transformer}, concatenated with the pre-trained Word2Vec word embeddings of \citet{MikolovEtAl-2013-Word2vec}, $D=4396$. The BERT model is pre-trained with masked-language modeling and next-sentence prediction objectives with large amounts of unlabeled data from 3.3 billion words. BERT's contextualized embeddings are capable of modeling dependencies between words and position information. The CNN can be viewed as summarizing the signal from this deep network for the fine-tuned task. We use the pre-trained, 340-million-parameter BERT\textsubscript{LARGE} model with case-preserving WordPiece \cite{WuEtAl-2016-GoogleNMT-WordPieceRef} tokenization.\footnote{We use the PyTorch (\url{https://pytorch.org/}) reimplementation of the original code base available at \url{https://github.com/huggingface/pytorch-pretrained-BERT} \cite{WolfEtAl-2020-HuggingFaceTransformers}.} In our experiments, we fine-tune the 300-dimensional word-embeddings with the CNN parameters, while the parameters of the BERT\textsubscript{LARGE} model remain fixed. The BERT model takes as input WordPiece tokens, using its full vocabulary, and we limit the vocabulary size to 7,500 only for the fine-tuned word embeddings. Prior to evaluation, to maintain alignment with the original tokenization and labels, the WordPiece tokenization is reversed (i.e., de-tokenized), with positive/negative token contribution scores averaged over fragments for original tokens split into separate WordPieces. We also consider fine-tuning the trained \textsc{uniCNN+BERT} model with the min-max loss, which we label \textsc{uniCNN+BERT+mm}.

\paragraph{Reference Models} We also include a reference base model, \textsc{cnn}, with filter widths of 3, 4, and 5, with 100 filter maps each, fine-tuning 300 dimensional Glove embeddings \cite{PenningtonEtAl-2014-Glove}, with a vocabulary of size 7,500, comparable to early work on zero-shot detection with lower parameter models. We additionally consider a model, \textsc{CNN+BERT}, similar to the primary \textsc{uniCNN+BERT} model, which uses Word2Vec word embeddings for consistency with the past supervised detection work of \citet{ReiAndYannakoudakis-2016-NeuralBaselines}, but with $Q$ and $M$ identical to \textsc{cnn}.  

\paragraph{Optimization and Tuning} For our zero-shot detection models, \textsc{cnn}, \textsc{CNN+BERT} and \textsc{uniCNN+BERT}, we optimize for sentence-level classification, choosing the training epoch with the highest sentence-level $F_1$ score on the dev set, without regard to token-level labels. These models do not have access to token-level labels for training or tuning.

We set aside 1k token-labeled sentences from the dev set to tune the token-level $F_{0.5}$ score for comparison purposes for the experiments labeled \textsc{CNN+BERT+1k} and \textsc{uniCNN+BERT+1k}.

\paragraph{\textsc{uniCNN+BERT+S*} Model}  
We also fine-tune a model with token-level labels, \textsc{uniCNN+BERT+S*}, with weights initialized with those of the \textsc{uniCNN+BERT} model trained for binary sentence-level classification. For calculating the loss at training, we assign each WordPiece to have the detection label of its original corresponding token, with the loss of a mini-batch averaged across all of the WordPieces. Inference is performed as in the zero-shot setting.\\

All models use dropout, with a probability of 0.5, applied on the output of the max-pooling operation, and we train with Adadelta \citep{Zeiler-2012-Adadelta} with a batch size of 50.

\subsubsection{Exemplar Auditing Decision Rules}
\paragraph{In-Domain Data}

For each of the \textsc{uniCNN} models, we also evaluate using the inference-time decision rules of Section~\ref{sec:methods-exa}, which we indicate with \textsc{+ExAG} and \textsc{+ExAT} appended to the model labels. The Euclidean distances are calculated at the word level of the original sentences, where we average the exemplar vectors when a word is split across multiple WordPiece tokens.
%\textsc{uniCNN+BERT+S*+ExA},\textsc{uniCNN+BERT+ExA},\textsc{uniCNN+BERT+mm+ExA}

\paragraph{Expanded Database with Domain-shifted Data}

We also consider adding the \textsc{FCE+news50k} data to the support set, and evaluating on the augmented \textsc{FCE+news2k} test set. For reference, we also \textit{train} the primary zero-shot models using the \textsc{FCE+news50k} data, for which we use the labels \textsc{uniCNN+BERT+news50k} and \textsc{uniCNN+BERT+mm+news50k}.

\subsubsection{K-NN Approximations}

We train each of the 3 proposed K-NN approximations on the held-out \textsc{KNN dev} set to minimize $\delta^{KNN}$, for up to 40 epochs, only using the predictions from the original models, rather than ground-truth labels. Only for the fully-supervised model, \textsc{uniCNN+BERT+S*} , do we then subsequently use token-level labels to tune the decision boundary, as with that original model. We add the labels of Section~\ref{sec:methods-k-nn} as suffixes to the original models to indicate the type of K-NN used, \textsc{+K$_8$NN\textsubscript{dist.}}, \textsc{+K$_8$NN\textsubscript{const.}}, \textsc{+K$_8$NN\textsubscript{equal}}, with the subscript indicating $K=8$. We chose $K=8$ on the held-out dev set based on minimizing $\delta^{KNN}$ with the \textsc{uniCNN+BERT+mm} model with $K \in \{1,3,5,8,25\}$. The approximations are only marginally better with $K=25$ for some of the models, so we hold $K=8$ constant for comparison purposes, and since smaller values of $K$ are preferable for interpretability, ceteris paribus. For reference, we also include results with K$_1$NN\textsubscript{equal}, which only considers the nearest match.  

\paragraph{Constraints for Domain-shifted Data} We also demonstrate constraining the output based on the maximum allowed distance to the nearest match in the support set, among matches for which the K-NN prediction equals that of the sentence-level label of the nearest match, and/or limited to minimum output magnitudes of the K-NN. We determine these constraints on the \textsc{KNN dev} set, based on $\delta^{KNN}$, determined without access to token-level labels; for simplicity, we use the mean values among correct \textit{approximations}. We examine this with weak models over the \textsc{FCE+news2k} domain-shifted test set that only have the FCE training set in the support set, investigating whether we can nonetheless identify subsets with strong effectiveness. This is a challenging but very practical setting, as in real deployments, the input data will often diverge from what we have seen in training. Such constraints serve as heuristics, tied to the model itself, for determining when to refrain from predicting, as is critical in higher-risk settings.\footnote{Within the set of admitted predictions, we might then consider approaches for quantifying uncertainty, which we leave for future work. Here we focus on examining and establishing the K-NN behavior relative to the original model to justify its use as an interpretable substitute, as well as the types of interpretable heuristics useful for avoiding domain-shifted and out-of-domain data this enables.}

\subsubsection{Previous Approaches and Baselines}

\paragraph{Previous Zero-shot Sequence Models} Recent work has approached zero-shot error detection by modifying and analyzing bidirectional LSTM taggers, which have been shown to work comparatively well on the task in the supervised setting. \citet{ReiAndSogaard-2018-ZeroShotSeq} adds a soft-attention mechanism to a bidirectional LSTM tagger, training with additional loss functions to encourage the attention weights to yield more accurate token-level labels (\textsc{LSTM-ATTN-SW}). Previous work also considered a gradient-based approach to analyze this same model (\textsc{LSTM-ATTN-BP}) and the model without the attention mechanism (\textsc{LSTM-LAST-BP}), by fitting a parametric Gaussian model to the distribution of magnitudes of the gradients of the word representations. % with respect to the sentence-level loss.

\paragraph{Previous Supervised Sequence Models} For comparison, we include recent fully supervised sequence models. \citet{ReiAndYannakoudakis-2016-NeuralBaselines} compares various word-based neural sequence models, finding that a word-based bidirectional LSTM model was the most effective (\textsc{LSTM-BASE+S*}). \citet{ReiAndSogaard-2018-ZeroShotSeq} compares against a bidirectional LSTM tagger with character representations concatenated with word embeddings (\textsc{LSTM+S*}). The model of \citet{Rei-2017-DetectionWithLMAuxLoss} extends this with an auxiliary language modeling objective (\textsc{LSTM+LM+S*}). This model is further enhanced with a character-level language modeling objective and supervised attention mechanisms in \citet{ReiAndSogaard-2018-SupervisedAttnTagging} (\textsc{LSTM+JOINT+S*}). \citet{BellEtAl-2019-LSTM+BERT-grammar-detection} considers BERT embeddings with the \textsc{LSTM+LM+S*} model, establishing a new state-of-the-art for the supervised setting, using a frozen BERT\textsubscript{BASE} model (\textsc{LSTM+LM+BERT\textsubscript{BASE}+S*}), and also providing results with a BERT\textsubscript{LARGE} model (\textsc{LSTM+LM+BERT+S*}).

\paragraph{Additional Baselines}

For reference, we also provide a \textsc{Random} baseline, which classifies based on a fair coin flip, and a \textsc{MajorityClass} baseline, which in this case always chooses the positive (``error detected'') class.

%%%%%%%%%%%%
%%%% Results
%%%%%%%%%%%%
\subsection{Grammatical Error Detection: Results}

\subsubsection{Zero-shot Results}

\begin{table}
\caption[Main results]{FCE test set results. The \textsc{LSTM} model results are as reported in \citet{ReiAndSogaard-2018-ZeroShotSeq}. With the exception of \textsc{LSTM+S*}, all models only have access to sentence-level labels while training. The sentence-level $F_1$ scores for the CNN models are from the fully-connected layer and are provided for reference.} 
\label{table:test-results-zero}
%\centering
%\footnotesize
\begin{tabular}{lcccc}
\toprule
 & \multicolumn{1}{c}{Sent} & \multicolumn{3}{c}{Token-level} \\
Model & $F_1$ & P & R & $F_1$ \\
\midrule
\textsc{LSTM+S*} & - & 49.15 & 26.96 & 34.76 \\
\midrule
\textsc{Random} & 58.30 & 15.30 & 50.07 & 23.44 \\
\textsc{MajorityClass} & 80.88 & 15.20 & 100. & 26.39 \\
\midrule
\textsc{LSTM-LAST-BP} & 85.10 & 29.49 & 16.07 & 20.80 \\
\textsc{LSTM-ATTN-BP} & 85.14 & 27.62 & 17.81 & 21.65 \\
\textsc{LSTM-ATTN-SW} & 85.14 & 28.04 & 29.91 & 28.27 \\
\midrule
\textsc{cnn} & 84.24 & 20.43 & 50.75 & 29.13 \\
\midrule
\textsc{CNN+BERT} & 86.35 & 26.76 & 61.82 & 37.36 \\
\textsc{uniCNN+BERT} & 86.28 & 47.67 & 36.70 & 41.47 \\
\bottomrule
\end{tabular}
\end{table} 

Table~\ref{table:test-results-zero} contains the main results with the models only given access to sentence-level labels, as well as \textsc{LSTM+S*} for reference, using $F_1$, as in previous zero-shot work. The task is very challenging, in general, with some baselines falling below random at the token level. The \textsc{cnn} model has a similar $F_1$ score as \textsc{LSTM-ATTN-SW}, and is stronger than the back-propagation-based approaches of \textsc{LSTM-ATTN-BP} and \textsc{LSTM-LAST-BP}. This is important, as it suggests the decomposition used with the basic \textsc{cnn} model, which amounts to a very lightweight attention mechanism, has the inductive bias suitable for such local detections, while being trivial to break apart into representative dense vectors of the input, enabling our analysis and interpretability methods. This is further confirmed when adding the pre-trained contextualized embeddings from BERT; remarkably, as a point of reference, these models exceed basic \textit{supervised} LSTM models that use pre-trained word embeddings. In Table~\ref{table:test-results-zeroAndSupervised} against $F_{0.5}$, which is the typical metric for evaluating supervised grammatical error detection, used under the assumption that end users prefer higher precision systems, the \textsc{uniCNN+BERT} model exceeds the fully supervised \textsc{LSTM-BASE+S*} model, which was the state-of-the-art model on the task as recently as 2016. 

Fine-tuning the zero-shot model \textsc{uniCNN+BERT} with the min-max loss constraint (\textsc{uniCNN+BERT+mm}) has the effect of increasing precision and decreasing recall, as seen in Table~\ref{table:test-results-zeroAndSupervised}. This results in a modest increase in $F_{0.5}$, but also a decrease in $F_1$ to 38.04. Whether or not this is a desirable tradeoff depends on the particular use case, but illustrates biasing the detections via task-specific constraints in the absence of token-level labels.

%The filter width over the deep network is an important inductive bias, with the \textsc{uniCNN+BERT} model substantially improving precision over the otherwise similar \textsc{CNN+BERT} model.

\textit{The inductive bias of the architecture is important for token-level detections: Models with similar sentence-level classification results can have significantly different token-level results.} For example, \textsc{CNN+BERT} and \textsc{uniCNN+BERT} have similar sentence-level $F_1$ scores of around 86, despite differing token-level effectiveness, and the LSTM baselines all exhibit similar sentence-level $F_1$ scores yet have significantly different token-level scores.  As such, attention-style approaches are useful, but not sufficient, for analyzing model predictions over the non-identifiable parameters of deep models, further justifying the need for the proposed methods establishing auditable mappings to the support set.

\subsubsection{Supervised \& Dev-set-tuned Results}

\begin{table}
\caption[Additional results]{Comparisons with recent state-of-the-art supervised detection models on the FCE test set. Models marked with \textsc{+S*} have access to approximately 28.7k token-level labeled sentences for training and 2.2k for tuning. Models marked with \textsc{+1k} have access to 28.7k sentence-level labeled sentences for training and 1k token-level labeled sentences for tuning. The \textsc{uniCNN+BERT} and \textsc{uniCNN+BERT+mm} models only have access to sentence-level labeled sentences. The results of the \textsc{LSTM} models are as previously reported in the literature.} 
\label{table:test-results-zeroAndSupervised}
%\centering
%\footnotesize
%small
\begin{tabular}{lccc}
\toprule
 & \multicolumn{3}{c}{Token-level} \\
Model & P & R & $F_{0.5}$ \\
\midrule
\textsc{LSTM+JOINT+S*} & 65.53 & 28.61 & 52.07 \\
\textsc{LSTM+LM+S*} & 58.88 & 28.92 & 48.48 \\
\textsc{LSTM-BASE+S*} & 46.1 & 28.5 & 41.1 \\
\midrule
\textsc{LSTM+LM+BERT\textsubscript{BASE}+S*}  & 64.96  & 38.89 & 57.28 \\
\textsc{LSTM+LM+BERT+S*} & 64.51 & 38.79 & 56.96 \\
\midrule
\textsc{uniCNN+BERT+S*} & 75.00 & 31.40 & 58.70 \\
\midrule
\textsc{CNN+BERT+1k} & 47.11 & 28.83 & 41.81 \\ 
\textsc{uniCNN+BERT+1k} & 63.89 & 23.27 & 47.36 \\
\midrule
\textsc{uniCNN+BERT} & 47.67 & 36.70 & 44.98 \\
\textsc{uniCNN+BERT+mm} & 54.87 & 29.10 & 46.62 \\
\bottomrule
\end{tabular}
\end{table} 

Table~\ref{table:test-results-zeroAndSupervised} also compares dev-set-tuned and fully supervised models. For illustrative purposes, \textsc{CNN+BERT+1k} and \textsc{uniCNN+BERT+1k} are given access to 1,000 token-labeled sentences to tune a single parameter, an offset on the decision boundary, for each model. This yields modest gains for both models, but interestingly, the \textsc{uniCNN+BERT}, in particular, already has a strong $F_{0.5}$ score without modification of the decision boundary in the true zero-shot setting.   

The \textsc{uniCNN+BERT+S*} model is a strong supervised sequence labeler. As seen in Table~\ref{table:test-results-zeroAndSupervised}, it is nominally stronger than the current state-of-the-art models recently presented in \citet{BellEtAl-2019-LSTM+BERT-grammar-detection}. This is critical, as it suggests we can forgo more complicated, expressive final layers, and instead use our proposed CNN and linear decomposition to, in effect, summarize the signal from the deep network, from which it is then straightforward to yield representations for matching, as analyzed next.

\subsubsection{Inference-time Decision Rules \& K-NN Approximations}

\begin{table}
\caption[Exemplar auditing results]{FCE test set results with the inference-time decision rules and replacing the original model with a K-NN approximation.}

\label{table:test-results-zeroAndSupervisedExemplarAuditing}
%\centering
%\footnotesize
\begin{tabular}{lccc}
\toprule
 & \multicolumn{3}{c}{Token-level} \\
Model & P & R & $F_{0.5}$ \\
\midrule
%\textsc{uniCNN+BERT+S*+ExA} & 85.14 & 22.00 & 54.09 \\
\textsc{uniCNN+BERT+S*+ExAG} & 85.17 & 21.86 & 53.93 \\
\textsc{uniCNN+BERT+S*+K$_1$NN\textsubscript{equal}} & 72.64 & 25.52 & 53.05 \\
\textsc{uniCNN+BERT+S*+K$_8$NN\textsubscript{dist.}} & 71.91  &  32.24 & 57.71 \\
\midrule
%\textsc{uniCNN+BERT+ExA} & 56.36 & 26.92 & 46.24 \\
\textsc{uniCNN+BERT+ExAG} & 56.79 & 26.74 & 46.37 \\
\textsc{uniCNN+BERT+K$_1$NN\textsubscript{equal}} & 47.23 & 32.01 & 43.13 \\
\textsc{uniCNN+BERT+K$_8$NN\textsubscript{dist.}} & 51.19 &  35.53 & 47.04 \\
\midrule
%\textsc{uniCNN+BERT+mm+ExA} & 63.79 & 20.03 & 44.39 \\
\textsc{uniCNN+BERT+mm+ExAG} & 63.88 & 20.03  & 44.43  \\
\textsc{uniCNN+BERT+mm+K$_1$NN\textsubscript{equal}} & 60.76 & 21.17 & 44.23 \\
\textsc{uniCNN+BERT+mm+K$_8$NN\textsubscript{dist.}} & 62.06 &  25.38 & 48.14 \\
\bottomrule
\end{tabular}
\end{table}

\begin{table}
\caption[KNN dev sign flips]{Additional results on the K-NN held-out dev set. $F_{0.5}$ and accuracy of the approximation ($\hat y^{KNN}=\hat y$), and $F_{0.5}$ of the K-NN against ground-truth ($\hat y^{KNN}=y$). The effectiveness of the original models ($\hat y=y$) on this subset of 14,867 tokens from 972 sentences is included for reference. } 
\label{table:fce-dev-knn-results}
%\centering
%\footnotesize
\small
\begin{tabular}{lccc}
\toprule
% & \multicolumn{3}{c}{Token-level} \\
& \multicolumn{1}{c}{True Labels} & \multicolumn{2}{c}{Model Approx.}\\
& \multicolumn{1}{c}{$\hat y^{KNN}=y$} & \multicolumn{2}{c}{$\hat y^{KNN}=\hat y$}\\
\cmidrule[0.75pt](lr){2-2} \cmidrule[0.75pt](lr){3-4}\\
Model & $F_{0.5}$ & Accuracy & $F_{0.5}$\\
\midrule
\textsc{uniCNN+BERT+S*+K$_1$NN\textsubscript{equal}} & 56.5 & 96.5 & 72.5\\
\textsc{uniCNN+BERT+S*+K$_8$NN\textsubscript{equal}} & 58.1 & 96.9 & 75.9\\
\textsc{uniCNN+BERT+S*+K$_8$NN\textsubscript{const.}} & 60.0 & 97.0 & 75.8\\
\textsc{uniCNN+BERT+S*+K$_8$NN\textsubscript{dist.}} & 59.4  & 97.0 & 75.9 \\
\midrule
\textsc{uniCNN+BERT+K$_1$NN\textsubscript{equal}} & 45.1 & 92.8 & 69.1\\
\textsc{uniCNN+BERT+K$_8$NN\textsubscript{equal}} & 50.5 & 94.2 & 78.0\\
\textsc{uniCNN+BERT+K$_8$NN\textsubscript{const.}} & 47.5 & 94.2 & 75.5 \\
\textsc{uniCNN+BERT+K$_8$NN\textsubscript{dist.}} & 48.1 & 94.3 & 76.4 \\
\midrule
\textsc{uniCNN+BERT+mm+K$_1$NN\textsubscript{equal}} & 47.7 & 95.8 & 72.4\\
\textsc{uniCNN+BERT+mm+K$_8$NN\textsubscript{equal}} & 52.3 & 96.4 & 76.9\\
\textsc{uniCNN+BERT+mm+K$_8$NN\textsubscript{const.}} &  53.1 & 96.4 & 75.9\\
\textsc{uniCNN+BERT+mm+K$_8$NN\textsubscript{dist.}} & 52.9 & 96.5 & 76.9\\
\midrule
& \multicolumn{1}{c}{True Labels} & \\
& \multicolumn{1}{c}{$\hat y=y$} & \\
\cmidrule[0.75pt](lr){2-2} \\
Model & $F_{0.5}$ &  & \\
\midrule
\textsc{uniCNN+BERT+S*} & 59.5 & - & -\\
\textsc{uniCNN+BERT} & 44.9 & - & -\\
\textsc{uniCNN+BERT+mm} & 49.6 & - & -\\
\bottomrule
\end{tabular}
\end{table}

\begin{figure}[hbt!]
\centering
\begin{subfigure}{.33\textwidth}
  \centering
  \includegraphics[width=\linewidth]{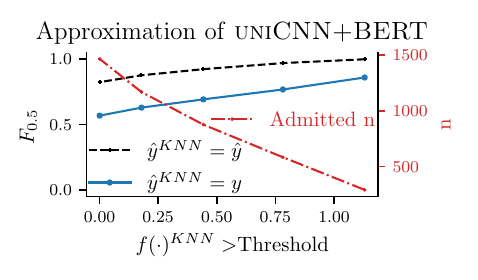}

\end{subfigure}%
\begin{subfigure}{.33\textwidth}
  \centering
  \includegraphics[width=\linewidth]{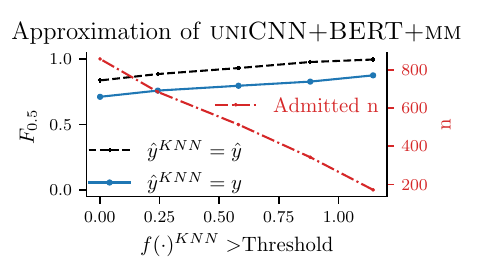}

\end{subfigure}%
\begin{subfigure}{.33\textwidth}
  \centering
  \includegraphics[width=\linewidth]{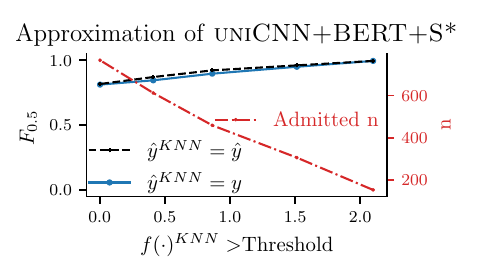}
\end{subfigure}%

\begin{subfigure}{.33\textwidth}
  \centering
  \includegraphics[width=\linewidth]{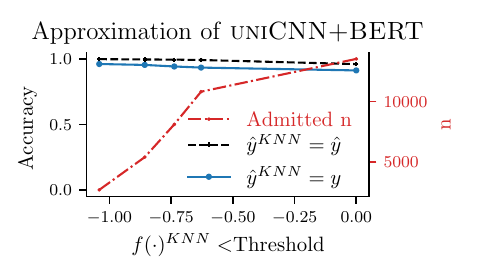}

\end{subfigure}%
\begin{subfigure}{.33\textwidth}
  \centering
  \includegraphics[width=\linewidth]{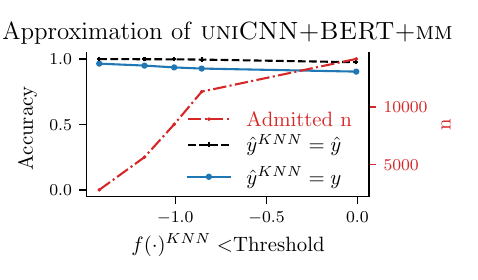}

\end{subfigure}%
\begin{subfigure}{.33\textwidth}
  \centering
  \includegraphics[width=\linewidth]{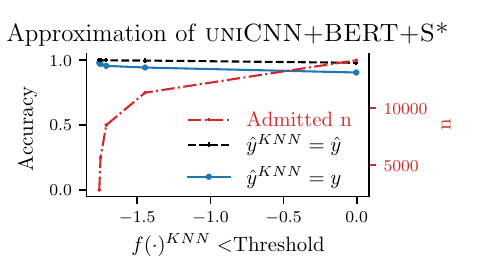}
\end{subfigure}%
\caption{On the in-domain K-NN dev split, across models, for both $\hat{y}^{KNN}_n=1$ (top row) and $\hat{y}^{KNN}_n=-1$ (bottom row), the $F_{0.5}$ and accuracy scores of the approximation (black dotted lines) generally track those of the K-NN against the ground-truth (blue lines) as the magnitude of the K-NN output varies. I.e., both the approximation and the prediction effectiveness increase with greater output magnitudes.} %Note the varying $n$ (right vertical axis) across models.}
\label{fig:fceknndev-approximation}
\end{figure}

\paragraph{In-domain Data}

Table~\ref{table:test-results-zeroAndSupervisedExemplarAuditing} shows the proposed exemplar auditing decision rules and the K-NN approximations on in-domain data across models. Compared to the results in Table~\ref{table:test-results-zeroAndSupervised}, the \textsc{ExAG} rule increases precision. In practice, matches tend to correspond to similar contexts, at least when the distance to the nearest exemplar in the support set is low, as shown in the examples in Appendix B. Further, the $F_{0.5}$ scores suggest that with $K=8$, the distance-weighted K-NNs (\textsc{KNN\textsubscript{dist.}}) are \textit{sufficient for replacing the original models' predictions}: The zero-shot K-NNs are nominally stronger than the corresponding original models, and the supervised version has the same effectiveness as the original for all practical purposes ($\pm1$ point). Note, too, that the precision vs. recall patterns for \textsc{uniCNN+BERT+mm+K$_8$NN\textsubscript{dist.}} vs. \textsc{uniCNN+BERT+K$_8$NN\textsubscript{dist.}} parallel those of \textsc{uniCNN+BERT+mm} vs. \textsc{uniCNN+BERT}, reflecting that the approximations are reasonably similar to the original models' predictions, especially over the subset of data for which the original models' predictions are correct, as discussed below.

We further examine the K-NN behavior on the held-out dev set in Table~\ref{table:fce-dev-knn-results}. We find that with $K=8$, across models, each of the proposed K-NN formulations can be trained to be roughly similar in approximation effectiveness, and when we reveal the true labels, there is not a clear  winner. In this way, the modeling choice shifts to other aspects of the model: The relative distances within the top-$K$ appear not to be critical on this dataset and can be replaced with constant learned weights with \textsc{KNN\textsubscript{const.}}; however, that comes at the expense of additional parameters and is harder to train due to the sensitivity of parameter initialization. The simplicity of \textsc{KNN\textsubscript{equal}} is appealing, but \textsc{KNN\textsubscript{dist.}} provides an explicit ranking over the exemplars with the addition of just a single learned parameter, so we take it as our primary model.

 As shown in Figure~\ref{fig:fceknndev-approximation}, across both classes and all models, the approximation effectiveness \textit{and} the K-NN's prediction effectiveness increase as the magnitude of the K-NN's output increases. This reflects a more general pattern: When the original model and/or K-NN produce incorrect predictions, the original model and the K-NN are more likely to produce different predictions. Put another way, \textit{difficult instances to predict also tend to be difficult instances over which to approximate the model}, which we can exploit as a heuristic to abstain from predicting, discussed below.

\begin{table}
\caption[Exemplar auditing results on out-of-domain data]{Domain-shifted \textsc{FCE+news2k} test set. The training set and the support set, $\sS$, differ in whether they include the FCE training set (F) or the \textsc{FCE+news50k} set (F+50k), or those sets with token-level labels (F* and F*+50k*, respectively).} 
\label{table:test-results-zeroAndSupervisedExemplarAuditingOutOfDomainWithGigaword}
%\centering
%\footnotesize
%\scriptsize
\small
\begin{tabular}{lccccc}
\toprule
 &  & &  \multicolumn{3}{c}{Token-level} \\
Model & Training & $\sS$ & P & R & $F_{0.5}$ \\
\midrule
\textsc{uniCNN+BERT+S*} & F* & - & 43.44 & 31.42 & 40.35 \\
\textsc{uniCNN+BERT+S*+ExAT} & F* & F* & 59.23 & 21.02 & 43.43 \\
\textsc{uniCNN+BERT+S*+ExAT} & F* & F*+50k* & 83.31 & 18.92 & 49.57 \\
\textsc{uniCNN+BERT+S*+K$_8$NN\textsubscript{dist.}} & F* & F* & 43.98 & 32.23 & 40.99 \\
\textsc{uniCNN+BERT+S*+K$_8$NN\textsubscript{dist.}} & F* & F*+50k* & 65.39 & 29.58 &  52.64 \\        
\midrule
\textsc{uniCNN+BERT+news50k} & F+50k & - & 26.64 & 40.13 & 28.56 \\
%\textsc{uniCNN+BERT+news50k+ExA50k} & & & 36.71 & 28.72 & 34.78 \\
\textsc{uniCNN+BERT+news50k+ExAG} & F+50k & F+50k & 47.10 & 26.55 & 40.79 \\
%\midrule
\textsc{uniCNN+BERT+mm+news50k} & F+50k & -  & 61.80 & 11.67 & 33.25 \\
%\textsc{uniCNN+BERT+mm+news50k+ExA50k} & & & 69.00 & 06.42 & 23.40 \\
\textsc{uniCNN+BERT+mm+news50k+ExAG} & F+50k & F+50k & 68.89 & 06.39 & 23.31 \\
\midrule
\textsc{uniCNN+BERT} & F & - & 21.84 & 36.65 & 23.76 \\        
\textsc{uniCNN+BERT+ExAG} & F & F & 29.19 & 26.74 & 28.66 \\ 
\textsc{uniCNN+BERT+ExAG} & F & F+50k & 56.39 & 23.52 & 44.07 \\ 
\textsc{uniCNN+BERT+ExAT} & F & F*+50k* & 75.98 & 18.51 & 46.87 \\ 
\textsc{uniCNN+BERT+K$_8$NN\textsubscript{dist.}} & F & F & 24.65 & 35.54 & 26.26 \\  
\textsc{uniCNN+BERT+K$_8$NN\textsubscript{dist.}} & F & F+50k & 43.64 & 30.91 & 40.32 \\   
\midrule
\textsc{uniCNN+BERT+mm} & F & - & 25.04 & 29.10 & 25.76 \\
%\textsc{uniCNN+BERT+mm+ExA} & & & 31.24 & 20.03 & 28.10 \\ 
\textsc{uniCNN+BERT+mm+ExAG} & F & F & 31.29 & 20.03 & 28.13 \\ % Not a typo. Note that if we apply the ground-truth constraint, there is essentially no change since it is only seeing the standard exemplar database with the FCE training set (and the model is relatively strong already)
\textsc{uniCNN+BERT+mm+ExAG} & F & F+50k & 65.08 & 17.62 & 42.30 \\
\textsc{uniCNN+BERT+mm+ExAT} & F & F*+50k* & 78.16 & 14.53 & 41.66 \\
%\textsc{uniCNN+BERT+mm+ExA50k} & & & 30.76 & 19.94 & 27.75 \\
\textsc{uniCNN+BERT+mm+K$_8$NN\textsubscript{dist.}} & F & F & 27.41 & 25.38 & 26.98 \\  
\textsc{uniCNN+BERT+mm+K$_8$NN\textsubscript{dist.}} & F & F+50k & 64.48 & 21.71 & 46.26 \\         
\bottomrule
\end{tabular}
\end{table}

\begin{table}
\caption[Distance constraints]{\textsc{FCE+news2k} test set. The output is constrained by a maximum allowed distance to the nearest match in the support set, among matches for which the K-NN prediction equals that of the sentence-level label of the nearest match, and/or limited to minimum output magnitudes of the K-NN. Constraints and thresholds are the mean values among correct \textit{approximations} on the K-NN dev set, determined without access to token-level labels. These limits identify subsets with significantly increased $F_{0.5}$ (cf., Table~\ref{table:test-results-zeroAndSupervisedExemplarAuditingOutOfDomainWithGigaword}), at the expense of not producing predictions for tokens over which the model is less reliable. Only 41,477 (out of $N=92,597$) of the tokens in this set are from the original FCE in-domain sentences. Without thresholds, the decision boundary is 0.} 
\label{table:testnews2k-results-constraints}
%\centering
\footnotesize
%\small
\begin{tabular}{lcccccc}
\toprule
 &  &  \multicolumn{1}{c}{$L^2$ distance} & \multicolumn{1}{c}{Output} &  & & \\
 &  & max constraint & min threshold &  Admitted & & \\
Model & $F_{0.5}$ & (Class -1, Class 1) & (Class -1, Class 1) &  $n$ & $n/N$\\
\midrule
\textsc{uniCNN+BERT+S*+K$_8$NN\textsubscript{dist.}} & 42.5 &  & & 92597 & 1.0\\
\textsc{uniCNN+BERT+S*+K$_8$NN\textsubscript{dist.}} & 62.5 &  & (-1.6, 1.3) & 53396 & 0.58\\
\textsc{uniCNN+BERT+S*+K$_8$NN\textsubscript{dist.}} & 67.5 & (25.3, 38.9) &  & 7896 & 0.09 \\
\textsc{uniCNN+BERT+S*+K$_8$NN\textsubscript{dist.}} & 86.9 & (25.3, 38.9) & (-1.6, 1.3) & 4219 & 0.05\\
\midrule
\textsc{uniCNN+BERT+K$_8$NN\textsubscript{dist.}} & 26.3 &  &  & 92597 & 1.0\\
\textsc{uniCNN+BERT+K$_8$NN\textsubscript{dist.}} & 46.5 &  & (-0.8, 0.7) & 40691 & 0.44\\
\textsc{uniCNN+BERT+K$_8$NN\textsubscript{dist.}} & 42.6 & (31.0, 47.6) &  & 8779 & 0.09\\
\textsc{uniCNN+BERT+K$_8$NN\textsubscript{dist.}} & 67.4 & (31.0, 47.6) & (-0.8, 0.7) & 4388 & 0.05\\
\midrule
\textsc{uniCNN+BERT+mm+K$_8$NN\textsubscript{dist.}} & 27.0 &  &  & 92597 & 1.0 \\
\textsc{uniCNN+BERT+mm+K$_8$NN\textsubscript{dist.}} & 45.9 &  & (-1.2, 0.8) & 38110 & 0.41 \\
\textsc{uniCNN+BERT+mm+K$_8$NN\textsubscript{dist.}} & 53.5 & (34.2,  53.3) &  & 7879 & 0.09 \\
\textsc{uniCNN+BERT+mm+K$_8$NN\textsubscript{dist.}} & 75.8 & (34.2,  53.3) & (-1.2, 0.8) & 4180 & 0.05 \\
\bottomrule
\end{tabular}
\end{table} 

%linear_exa.epoch15.database_train.query_test.tanh_knn8_fce_dev_split_temp10.0_model-maxent.fcenews2k.graph.base_f05.eps is fce2k_magnitude_sort_supervised_f05.eps
%linear_exa.epoch15.database_train.query_test.tanh_knn8_fce_dev_split_temp10.0_model-maxent.fcenews2k.graph.base_f05.eps is fce2k_magnitude_sort_base_f05.eps

\begin{figure}[hbt!]
\centering
\begin{subfigure}{.5\textwidth}
  \centering
  \includegraphics[width=\linewidth]{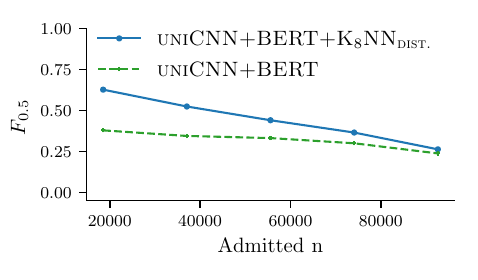}
\end{subfigure}%
\begin{subfigure}{.5\textwidth}
  \centering
  \includegraphics[width=\linewidth]{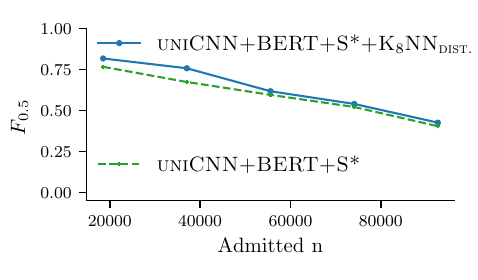}
\end{subfigure}
\caption{The original model output and the K-NN approximation output as comparative measures of prediction reliability on the domain-shifted \textsc{FCE+news2k} test set. The predictions are sorted by the magnitude of the output and scored in 5 bins. We consider both classes together, holding $n$ constant within bins. The magnitude of the K-NN output tracks prediction reliability at least as well as that of the original model, with the advantage that the K-NN has an explicit, interpretable connection to the support set and available labels. Appendix Table~\ref{table:fce-news2k-domain-shifted-test-magnitude-quantiles} similarly examines \textsc{uniCNN+BERT+mm+K$_8$NN\textsubscript{dist.}} looking at each class separately.}
\label{fig:fcenews2k-test-magnitude-comparator}
\end{figure}

\paragraph{Domain-shifted Data} Table~\ref{table:test-results-zeroAndSupervisedExemplarAuditingOutOfDomainWithGigaword} considers the more challenging setting in which the FCE test set has been augmented with 2,000 already correct sentences in the news domain. Just applying the \textsc{uniCNN+BERT+mm} model to this modified test set yields a large number of false positives on the already correct data, yielding a $F_{0.5}$ of 25.76 (c.f., the $F_{0.5}$ score of 46.62 on the original test set, as shown in Table~\ref{table:test-results-zeroAndSupervised}), and similarly for the other models, including that with full supervision. Simply training with the domain-shifted data, as with \textsc{uniCNN+BERT+news50k}, still results in low effectiveness for the zero-shot models, presumably owing to the class imbalance. Furthermore, the $F_{0.5}$ score of the \textsc{uniCNN+BERT+news50k} model on the original FCE test set (a result not shown in the tables) is 39.57, which is lower than the result of 44.98 of \textsc{uniCNN+BERT}, the equivalent model trained only with the original FCE set (Table~\ref{table:test-results-zeroAndSupervised}). 

However, when we update the support set with the domain-shifted data, in conjunction with the decision rules or the K-NN approximations, the $F_{0.5}$ scores jump significantly across models. The models are generally weak predictors over the domain-shifted data, but the improved scores reflect the capacity of the representations to match to the new data, and by extension, the associated labels. This mechanism opens the potential to update the model locally without a full re-training.

Matching to the support set in this way can improve effectiveness over domain-shifted data, but of course, it also requires such data to be in the support set prior to inference. In practice, it may be advisable to include as much data in the support set as computationally feasible, refraining from predicting for matches to unlabeled data, as applicable. In higher-risk settings, we can also constrain predictions based on the $L^2$ distance to the nearest match and the magnitude of the K-NN output, as demonstrated in Table~\ref{table:testnews2k-results-constraints} on the \textsc{FCE+news2k} test set. These constraints limit predictions to reliable subsets, even for these models that are weak predictors over the full set. These heuristics are interpretable in that the matched distance can be compared to other instances, and the K-NN output is a bounded value that is an explicit weighting over instances with known labels, tracking prediction reliability at least as well as the magnitude of the token-level output of the original model (Figure~\ref{fig:fcenews2k-test-magnitude-comparator}). %, exploiting the aforementioned patterns with respect to the approximations for cases when token-level labels are unavailable

\subsection{Grammatical Error Detection: Discussion}
The baseline expectations for zero-shot grammatical error detection models are low given the difficulty of the supervised case. It is therefore relatively surprising that a model such as \textsc{uniCNN+BERT}, when given only sentence-level labels, can yield a reasonably decent sequence model that is in the ballpark of some recent---even if lower parameter---fully supervised models. The inductive bias of the proposed method over a strong deep network is effective for such class-conditional detection, as well as supervised labeling. The approach additionally enables dense representation matching against a support set with known labels, with both inference-time decision rules and particular K-NN approximations. In this way, we gain the ability to make updates to a model without re-training; to constrain predictions based on interpretable heuristics; and more generally, to recast the otherwise black-box predictions of the network as an explicit weighting over instances with known labels.

%%%%%%%%%%%%%%%%%%%%%%%%%%%%%%%%%%%%%%%%%%
%%%% SECTION
%%%%%%%%%%%%%%%%%%%%%%%%%%%%%%%%%%%%%%%%%%
\section{Sentiment Data: Binary Prediction of Polarity}\label{sec:sentiment}

We further analyze the behavior of updating the support set over domain-shifted data for the task of predicting sentiment features in IMDb movie reviews. We consider recent work that re-annotates document-level classification data with minimal, local revisions that change the class labels \cite{Kaushik-etal-2019-CounterfactualDataAnnotations,Gardner-etal-2020-ContrastSets}, from which we back-out token-level labels for evaluation. We use this existing data-oriented approach for robust classification for controlled tests of the internal validity of our approach. We observe an ability to adapt the models via matching as with the grammar experiments. Additionally, in this context, we find that robust prediction over new, unseen domains remains challenging, but simple token-level heuristics tied to the K-NN approximation are nonetheless at least reasonably effective at constraining predictions to reliable subsets, and for screening data unlike that seen in training. This provides further justification for methods, such as proposed here, with which we can analyze and curate the data under the current generation of deep networks. %In Section~\ref{sec:local-annotation-detection} we examine class-conditional characteristics of the annotations in these sentiment datasets.

\subsection{Sentiment Data: Experiments}\label{sec:sentiment-experiments}
We consider the task of predicting binary document-level sentiment in IMDb movie reviews. We analyze detection of sentiment features at the token level, treating it as a zero-shot sequence labeling task, and additionally provide document-level classification results when constraining the predictions based on the token-level heuristics.

\subsubsection{Data: IMDb Sentiment (Negative vs. Positive) with Local Re-edits}

We use the IMDb data of \citet{Kaushik-etal-2019-CounterfactualDataAnnotations}\footnote{Available at \url{https://github.com/acmi-lab/counterfactually-augmented-data}.}. This consists of movie reviews with negative sentiment ($Y=-1$) and positive sentiment ($Y=1$), including reviews from the original review site (original, or \textsc{Orig.}) and ``counterfactually-augmented'' revisions (\textsc{Rev.}), the latter of which were created by crowd-workers who annotated the original reviews with local, minimal changes that change the document-level label. For document/review-level sentiment, we follow the main splits of the original work and train on a sample of 3.4k original reviews, \textsc{Orig.} (3.4k), and the original reviews combined with their corresponding revisions, \textsc{Orig.+Rev.} (1.7k+1.7k). For experiments modifying the support set, we will also consider each of these halves separately, \textsc{Orig.} (1.7k) and \textsc{Rev.} (1.7k). For reference, we additionally train with the full set of original reviews, \textsc{Orig.} (19k), and the full set combined with the revisions, \textsc{Orig.+Rev.} (19k+1.7k). For evaluation, we consider the \textsc{Orig.} and \textsc{Rev.} test sets from previous work. %, where each document consists of a movie review typically containing several sentences; , which means parallel source-target pairs are included, the combination of which matches the size of the 3.4k set of only originals for comparison purposes

To control for the language distribution of the revisions, we also create a new set of disjoint source-target pairs for training by removing the corresponding original reviews and leaving the revisions. We then add in disjoint samples from the remaining full set of original reviews to fill out the remaining sample size. For the smaller set this results in a set of 3.4k reviews, \textsc{Orig.\textsubscript{DISJOINT}+Rev.} (1.7k+1.7k), the same size as the comparable parallel set. For the larger set, we simply remove any original reviews that match the original reviews paired with revised reviews, creating \textsc{Orig.\textsubscript{DISJOINT}+Rev.} (19k-1.7k+1.7k). %For the dev split, for each parallel source-target pair in the dev file, we randomly select one of either the original or revised reviews. The resulting dev set is half the size of the parallel file.

\paragraph{Sentiment Diffs for Token-Level Detection} We use the parallel original and revision data to create token-level feature labels. Treating positive reviews as the source, we deterministically generate source-target transduction diffs in the same manner as \citet{Schmaltz-etal-2017-Diff-Transduction}. We then assign the positive class ($y_n=1$) to tokens associated with diffs that transduce to documents for which $Y=1$, assigning all other tokens to the negative class ($y_n=-1$). We use a similar convention as the the FCE dataset in Section~\ref{sec:error-detection} with respect to insertions, deletions, and replacements. Table~\ref{table-sentiment-diffs-data-creation} provides an example.

\subsubsection{Data: IMDb Sentiment (Negative vs. Positive) with Contrast Sets} We additionally evaluate on the IMDb reviews of \citet{Gardner-etal-2020-ContrastSets}\footnote{Available at \url{https://github.com/allenai/contrast-sets/tree/master/IMDb}}, which are revised with local re-edits by professional researchers familiar with the task instead of by crowd-sourced workers. This test set (\textsc{Contrast}) corresponds to the same set of reviews in the test set of \citet{Kaushik-etal-2019-CounterfactualDataAnnotations}. We do not have a corresponding training set, nor do we use the corresponding dev set for tuning, so we consider all evaluation on this set to be a domain-shifted setting.

\subsubsection{Data: Out-of-domain Twitter Document-level Sentiment (Negative vs. Positive)}

Finally, we also evaluate on the test set of SemEval-2017 Task 4a \cite{Rosenthal-etal-2017-Semeval2017Task4}\footnote{Available at \url{http://alt.qcri.org/semeval2017/task4/data/uploads/semeval2017-task4-test.zip}}. This consists of Twitter messages, which are significantly different than the IMDb movie reviews in terms of the topics covered, the language distribution, and the length of the documents, so we consider this to be an out-of-domain setting. We follow the previous work of \citet{Kaushik-etal-2019-CounterfactualDataAnnotations} in evaluating the binary classification results with accuracy. We balance the test set, using equal numbers of negative and positive Tweets, and drop the third class (neutral) for consistency with the earlier work, resulting in 4,750 Twitter messages for evaluation. 

\subsubsection{Models} 

Our core model is the \textsc{uniCNN+BERT} model from Section~\ref{sec:error-detection}, with which we vary the training set and the data in the support set. The only differences from \textsc{uniCNN+BERT} in the grammar detection experiments is that we set the maximum length, by WordPiece, to 350 as in previous works, and we choose the training epoch (up to a max of 60 epochs) by the highest accuracy on the dev set.

We evaluate token-level predictions of sentiment diffs using the $F_{0.5}$ metric, as with grammatical error detection above. We vary whether the support set includes data from the \textsc{Orig.} and/or \textsc{Rev.} training sets, using the labels \textsc{+ExAG} and \textsc{+ExAT} from Section~\ref{sec:methods-exa} to identify the particular rules used. We also present results where we allow the models a small amount of data to tune the decision boundary for the token-level predictions. For consistency, we always use the dev set of the \textsc{Orig.} reviews subset, using the subscript \textsubscript{\textsc{+orig\_dev}} to indicate that the models have access to 245 sentences with token-level labels. This provides a point of comparison to the exemplar auditing decision rules.

\paragraph{K-NN} We train the distance-weighted K-NN approximation on the held-out \textsc{KNN dev} set to minimize $\delta^{KNN}$ as in Section~\ref{sec:error-detection}, but for up to 60 epochs, \textsc{uniCNN+BERT+K$_8$NN\textsubscript{dist.}}. The original model is trained on the \textsc{Orig.} (3.4k) data. For comparisons with experiments with the inference-time decision rules, the K-NN is trained with \textsc{Orig.} (1.7k) as the support set, using half of the \textsubscript{\textsc{+orig\_dev}} for setting the K-NN parameters and the other half as the held-out \textsc{KNN dev}. This is a relatively limited amount of data, but it is sufficient for training the 3 parameters of the K-NN to at least match the accuracy of the original model.

\paragraph{K-NN Token-level Constraints for Document-level Classification} The K-NN enables interpretable heuristics for constraining predictions to the most reliable subsets of the data. In Section~\ref{sec:error-detection}, we demonstrated this for token-level detection; here, we show how this idea can be applied toward document-level classification, as well. As with detection in Table~\ref{table:testnews2k-results-constraints}, token-level predictions are constrained by a maximum allowed distance to the nearest match in the support set and K-NN output magnitude limits derived from correct \textit{approximations} on the \textsc{KNN dev} set, determined without access to token-level labels. For both distances and magnitudes, we use the mean for each class among correct approximations. Using the full \textsubscript{\textsc{+orig\_dev}} set we then set limits on the proportion and/or range of admitted tokens per document required to admit the overall document-level classification from the original \textsc{uniCNN+BERT} model. To emulate a high-risk setting, we set the minimum threshold such that all admitted document-level predictions are correct on the dev set. We also optionally further require the total number of tokens admitted to be within $\pm$ 1 standard deviation from the mean of correct predictions to control for unexpected lengths.\footnote{This differs from a simple hard constraint on token input lengths. In principle, most Twitter messages could still be admitted by this model-dependent constraint, as the lower bound is around 5 tokens.}

\paragraph{Previous Approaches} Our primary focus in this section is holding the model architecture from Section~\ref{sec:error-detection} constant while changing the data subsets. For reference, we include the results of \citet{Kaushik-etal-2019-CounterfactualDataAnnotations}, which fine-tunes the \textsc{BERT\textsubscript{BASE}} uncased model with the standard final linear layer for classification, \textsc{BERT\textsubscript{BASE\textsubscript{uncased}}+FT}. For comparison, we then also train a model using this same Transformer as frozen input with uncased Glove embeddings, \textsc{uniCNN+BERT\textsubscript{BASE\textsubscript{uncased}}}, and also an analogous cased model with Word2Vec embeddings, \textsc{uniCNN+BERT\textsubscript{BASE}}. 

\subsection{Sentiment Data: Results}\label{sec:sentiment-results}

\paragraph{Document-level Classification} For context, Table~\ref{table:test-results-sentiment-review-level} shows the document-level accuracy of \textsc{uniCNN+BERT} when varying the training data, tested on the original (\textsc{Orig.}) and revised (\textsc{Rev.}) test sets. Training with \textsc{Orig.} vs. \textsc{Orig.+Rev.} reflects the same patterns seen in the experiments of \citet{Kaushik-etal-2019-CounterfactualDataAnnotations}; however, if we control for the language of the revised reviews by training with disjoint source-target pairs (\textsc{Orig.\textsubscript{DISJOINT}+Rev.}), the difference across test sets is more modest. For reference, we find that \textsc{uniCNN+BERT} is at least as effective as fine-tuning all parameters of the \textsc{BERT\textsubscript{BASE}} model, with the \textsc{uniCNN+BERT\textsubscript{BASE\textsubscript{uncased}}} variant within 2-3 points (Table~\ref{table:test-results-sentiment-review-level-bert-ft-comparison}).

\paragraph{Document-level Classification with Token-level Constraints}

Table~\ref{table:test-results-sentiment-review-level-constraints} shows review-level test accuracy with \textsc{uniCNN+BERT} trained on the \textsc{Orig.} (3.4k) data using \textsc{uniCNN+BERT+K$_8$NN\textsubscript{dist.}} to determine constraints. Token-level predictions are constrained by a maximum allowed distance to the nearest match in the support set and K-NN output magnitude limits derived from correct \textit{approximations} on the K-NN dev set, determined without access to token-level labels (as in Table~\ref{table:testnews2k-results-constraints}). The document-level predictions are then constrained by a minimum threshold ($\approx 10\%$) on the proportion of admitted tokens among all tokens in the document and optionally, an additional constraint on the allowed range of admitted tokens (between 5 and 15, which is $\pm$ 1 standard deviation from the mean), both determined from sentence-level labels on the dev set. 

These simple, understandable constraints derived from the token-level predictions are effective at restricting the model to the most reliable document-level predictions, including on dramatically different out-of-domain input (\textsc{SemEval-2017}). For the constraints with the original (\textsc{Orig.}) and revised (\textsc{Rev.}) test sets, the same 3 and 1 reviews, respectively, are missed with both constraint variants, which accounts for the nominally lower accuracy as a result of a smaller denominator, and notably, 1 review in each of these sets is incorrectly or ambiguously annotated in the ground-truth data. On average, only around 1 token is admitted per Tweet in the \textsc{SemEval-2017} data with the distance and magnitude constraints, so the hard token count constraints readily filter most such data for document-level predictions, which is desirable given the mis-match with the training data. In contrast, the orthogonal approach of seeking more robust predictions by including source-target pairs was not consistently beneficial, as shown in Table~\ref{table:test-results-sentiment-semeval2017}. %, despite the significant additional annotation costs.

\paragraph{Token-level Feature Detection} The token-level feature detections follow a similar pattern with regard to the training data sets as the document-level predictions, with gains observed with the locally re-edited data, and to a lesser extent, the disjoint sets, as shown in Table~\ref{table:test-results-sentiment-sequence-level} and the true zero-shot setting shown in the \textcolor{red}{\texttt{red}} and black rows of Table~\ref{table:test-results-sentiment-sequence-level-exa-rules}. The predictions from the K-NN are at least as effective as the original model. As with the error detection experiments, the inference-time decision rules can be used to make updates to the model without retraining (Table~\ref{table:test-results-sentiment-sequence-level-exa-rules}), which in some cases, results in $F_{0.5}$ scores approaching that of training on that same data. 

The observed patterns are analogous on the professionally annotated \textsc{Contrast} test set, as shown in Tables~\ref{table:test-results-sentiment-contrast-sequence-level} and \ref{table:test-results-contrast-sentiment-sequence-level-exa-rules}. A relatively modest amount of labeled data in the support set is sufficient for improving effectiveness in detecting the token-level sentiment features as seen in the rightmost column of Table~\ref{table:test-results-contrast-sentiment-sequence-level-exa-rules}.

\begin{table*}[hbt!] %[t]
\caption[Counterfactually-augmented Sentiment Prediction Review-Level]{Predicting \textit{sentiment} on the original (\textsc{Orig.}) and revised (\textsc{Rev.}) test sets at the review level, using \textsc{uniCNN+BERT}, varying the training data (rows).} 
\label{table:test-results-sentiment-review-level}
%\centering
\footnotesize
%\scriptsize
\begin{tabular}{lcc}
\toprule
& \multicolumn{2}{c}{Review-level Sentiment (Accuracy)} \\
Model Train. Data (Num. Reviews) & \textsc{Orig.} & \textsc{Rev.} \\
\midrule
\textsc{Random} & 50.2 & 49.8 \\
\midrule
\textsc{Orig.} (3.4k) & 92.8 & 88.7 \\
\textsc{Orig.+Rev.} (1.7k+1.7k) & 90.6 & 96.5 \\
\textsc{Orig.\textsubscript{DISJOINT}+Rev.} (1.7k+1.7k) & 89.5 & 95.7 \\
\midrule
\textsc{Orig.} (19k) & 93.0 & 87.9 \\
\textsc{Orig.+Rev.} (19k+1.7k) & 93.0 & 94.3 \\
\textsc{Orig.\textsubscript{DISJOINT}+Rev.} (19k-1.7k+1.7k) & 93.0 & 90.2 \\
\bottomrule
\end{tabular}
\end{table*}

\begin{table*}[hbt!] %[t]
\caption[Counterfactually-augmented Sentiment Prediction Review-Level Constraints]{Review-level test accuracy with \textsc{uniCNN+BERT} trained on the \textsc{Orig.} (3.4k) data using \textsc{uniCNN+BERT+K$_8$NN\textsubscript{dist.}} to constrain predictions. Token-level predictions are constrained by a maximum allowed distance to the nearest match in the support set and K-NN output magnitude limits derived from correct approximations on the K-NN dev set. Document-level  predictions are admitted based on a minimum threshold on the proportion of admitted tokens among all tokens in the document (``Admitted token \% min'') and optionally, an additional constraint on the allowed range of admitted tokens (``Admitted token min, max'').} 
\label{table:test-results-sentiment-review-level-constraints}
%\centering
\footnotesize
%\scriptsize
\begin{tabular}{lccc cc}
\toprule
& \multicolumn{1}{c}{Review-level}  &  \multicolumn{1}{c}{Admitted} &  \multicolumn{1}{c}{Admitted}\\
& \multicolumn{1}{c}{Sentiment} &  \multicolumn{1}{c}{token \%} &  \multicolumn{1}{c}{token} &  Admitted & \\
Test Data & \multicolumn{1}{c}{(Accuracy)}  &  \multicolumn{1}{c}{min} &  \multicolumn{1}{c}{min, max} &  $n$ & $n/N$\\
\midrule
\textsc{Orig.} & 92.8  &  &  & 488 & 1.0\\
\textsc{Orig.} & 96.2 & $\bullet$ &  & 78 & 0.16\\
\textsc{Orig.} & 93.0  & $\bullet$ & $\bullet$ & 43 & 0.09\\
\midrule
\textsc{Rev.} & 88.7  &  &  & 488 & 1.0\\
\textsc{Rev.} & 98.1 & $\bullet$ &  & 52 & 0.11\\
\textsc{Rev.} &  97.0 & $\bullet$ & $\bullet$ & 33 & 0.07\\
\midrule
\textsc{SemEval-2017} & 77.8  &  &  & 4750 & 1.0\\
\textsc{SemEval-2017} & 81.4 & $\bullet$ &  & 576 & 0.12\\
\textsc{SemEval-2017} &  100.0 & $\bullet$ & $\bullet$ & 1 & 0.0002\\
\bottomrule
\end{tabular}
\end{table*}

\begin{table*}[hbt!] %[t]
\caption[Counterfactually-augmented Sentiment Prediction Seq.]{Predicting \textit{sentiment diffs} at the token level ($F_{0.5}$). All results are with the \textsc{uniCNN+BERT} model, varying the training data, except for the second row with \textsc{uniCNN+BERT+K$_8$NN\textsubscript{dist.}}. The decision boundary is tuned with token-level diffs from 245 \textsc{Orig.} dev set reviews (cf., the true zero-shot setting in Table~\ref{table:test-results-sentiment-sequence-level-exa-rules}).} 
\label{table:test-results-sentiment-sequence-level}
%\centering
\footnotesize
%\scriptsize
\begin{tabular}{lcc}
\toprule
& \multicolumn{2}{c}{Token-level Sentiment Diffs ($F_{0.5}$)}  \\
Model Train. Data (Num. Reviews) & \textsc{Orig.} & \textsc{Rev.} \\
\midrule
\textsc{Random} & 6.0 & 7.6 \\
\midrule
\textsc{Orig.} (3.4k)\textsubscript{\textsc{+orig\_dev}} (K$_8$NN\textsubscript{dist.}) & 29.5 & 23.5 \\ %\textsc{uniCNN+BERT+K$_8$NN\textsubscript{dist.}}
\midrule
\textsc{Orig.} (3.4k)\textsubscript{\textsc{+orig\_dev}} & 26.2 & 22.5 \\
\textsc{Orig.+Rev.} (1.7k+1.7k)\textsubscript{\textsc{+orig\_dev}} & 32.4 & 33.1 \\
\textsc{Orig.\textsubscript{DISJOINT}+Rev.} (1.7k+1.7k)\textsubscript{\textsc{+orig\_dev}} & 32.4 & 31.5 \\
\midrule
\textsc{Orig.} (19k)\textsubscript{\textsc{+orig\_dev}} & 24.8 & 21.7 \\
\textsc{Orig.+Rev.} (19k+1.7k)\textsubscript{\textsc{+orig\_dev}} & 28.8 & 27.9 \\
\textsc{Orig.\textsubscript{DISJOINT}+Rev.} (19k-1.7k+1.7k)\textsubscript{\textsc{+orig\_dev}} & 28.2 & 26.8 \\
\bottomrule
\end{tabular}
\end{table*} 

\begin{table*}[hbt!]
\caption[Counterfactually-augmented Sentiment Prediction Seq. ExA]{Predicting \textit{sentiment diffs} at the token level ($F_{0.5}$) with \textsc{uniCNN+BERT}, applying the exemplar auditing decision rules. Predictions without accessing the support set ($\sS$) are displayed in \textcolor{red}{\texttt{red}}. Underlined results indicate $\sS$ contains additional reviews or signal not seen by the model during training. Results with access to token-level labels in $\sS$ are further highlighted in \textcolor{blue}{\underline{blue}}.} 
\label{table:test-results-sentiment-sequence-level-exa-rules}
\centering
%\footnotesize
\scriptsize
\begin{tabular}{lcccc cc}
\toprule
& \multicolumn{6}{c}{Token-level Sentiment Diffs ($F_{0.5}$)}  \\
\cmidrule[0.75pt](lr){2-7}\\
%& \multicolumn{2}{c}{Database: \textsc{Orig.} (1.7k)} & \multicolumn{2}{c}{Database: \textsc{Rev.} (1.7k)} & \multicolumn{2}{c}{Database: \textsc{Orig.+Rev.} (1.7k+1.7k)} \\
& \multicolumn{2}{c}{$\sS=$} & \multicolumn{2}{c}{$\sS=$} & \multicolumn{2}{c}{$\sS=$} \\
& \multicolumn{2}{c}{\textsc{Orig.} (1.7k)} & \multicolumn{2}{c}{\textsc{Rev.} (1.7k)} & \multicolumn{2}{c}{\textsc{Orig.+Rev.} (1.7k+1.7k)} \\
\cmidrule[0.75pt](lr){2-3}\cmidrule[0.75pt](lr){4-5}\cmidrule[0.75pt](lr){6-7} \\
Model Train. Data (Num. Reviews) & Test: \textsc{Orig.} & Test: \textsc{Rev.} & Test: \textsc{Orig.} & Test: \textsc{Rev.} & Test: \textsc{Orig.} & Test: \textsc{Rev.}\\
\midrule
\textsc{Random} & 6.0 & 7.6 & 6.0 & 7.6 & 6.0 & 7.6 \\
\midrule
\textsc{Orig.} (3.4k) & \textcolor{red}{\texttt{10.5 }} & \textcolor{red}{\texttt{11.2 }} & \textcolor{red}{\texttt{10.5 }} & \textcolor{red}{\texttt{11.2 }} & \textcolor{red}{\texttt{10.5 }} & \textcolor{red}{\texttt{11.2 }} \\
%\textsc{Orig.+ExA} (3.4k) & 13.1 & 12.9 & \underline{13.1} &  \underline{12.8} & \underline{13.0} & \underline{12.9} \\
\textsc{Orig.+ExAG} (3.4k) & 16.9 & 15.8 & \underline{18.2} &  \underline{18.3} & \underline{18.0} & \underline{17.8} \\ 
\textsc{Orig.+ExAT} (3.4k) & \textcolor{blue}{\underline{34.7}}  & \textcolor{blue}{\underline{ 28.2}} &  \textcolor{blue}{\underline{30.1}} &  \textcolor{blue}{\underline{29.1}} & \textcolor{blue}{\underline{32.8}} & \textcolor{blue}{\underline{28.8}} \\ 
\midrule
\textsc{Orig.+Rev.} (1.7k+1.7k) & \textcolor{red}{\texttt{20.1}} & \textcolor{red}{\texttt{22.0}} & \textcolor{red}{\texttt{20.1}} & \textcolor{red}{\texttt{22.0}} & \textcolor{red}{\texttt{20.1}} & \textcolor{red}{\texttt{22.0}}\\
%\textsc{Orig.+Rev.+ExA} (1.7k+1.7k) & 27.4 & 28.4 & 27.0 & 28.4 & 27.0 & 28.8\\
\textsc{Orig.+Rev.+ExAG} (1.7k+1.7k) & 32.0 & 31.4 & 30.4 & 33.2 & 31.5 & 33.7 \\
\textsc{Orig.+Rev.+ExAT} (1.7k+1.7k) & \textcolor{blue}{\underline{39.2}} & \textcolor{blue}{\underline{33.3}} & \textcolor{blue}{\underline{31.6}} & \textcolor{blue}{\underline{36.1}} & \textcolor{blue}{\underline{37.4}} & \textcolor{blue}{\underline{37.1}}\\
\midrule
\textsc{Orig.\textsubscript{DISJOINT}+Rev.} (1.7k+1.7k) & \textcolor{red}{\texttt{15.8}} & \textcolor{red}{\texttt{17.4}} & \textcolor{red}{\texttt{15.8}} & \textcolor{red}{\texttt{17.4}} & \textcolor{red}{\texttt{15.8}} & \textcolor{red}{\texttt{17.4}}\\
%\textsc{Orig.\textsubscript{DISJOINT}+Rev.+ExA} (1.7k+1.7k) & \underline{22.8} & \underline{22.9} & 21.5 & 23.0 &  \underline{22.1} & \underline{22.6}\\
\textsc{Orig.\textsubscript{DISJOINT}+Rev.+ExAG} (1.7k+1.7k) & \underline{28.1} & \underline{26.8} & 26.0 & 28.3 &  \underline{27.3} &  \underline{28.1}\\
\textsc{Orig.\textsubscript{DISJOINT}+Rev.+ExAT} (1.7k+1.7k) & \textcolor{blue}{\underline{38.8}} & \textcolor{blue}{\underline{32.8}} & \textcolor{blue}{\underline{33.3}} & \textcolor{blue}{\underline{34.9}} &  \textcolor{blue}{\underline{37.0}} &  \textcolor{blue}{\underline{34.5}}\\
\bottomrule
\end{tabular}
\end{table*}

\begin{table*}[hbt!]
\caption[Zero-shot Contrast Set]{Predicting review-level sentiment (accuracy) and token-level \textit{sentiment diffs} ($F_{0.5}$) on the professionally annotated \textsc{Contrast} test set. In the second column, the decision boundary is the same as that tuned for Table~\ref{table:test-results-sentiment-sequence-level} using 245 \textsc{Orig.} dev set reviews, as indicated by the (\textsubscript{\textsc{+orig\_dev}}) label (cf., the true zero-shot setting in Table~\ref{table:test-results-contrast-sentiment-sequence-level-exa-rules}.)}
\label{table:test-results-sentiment-contrast-sequence-level}
%\centering
\footnotesize
%\scriptsize
\begin{tabular}{lcc}
\toprule
\multicolumn{3}{c}{\textbf{Contrast Sets}}  \\
\midrule
& \multicolumn{1}{c}{Review-level Sentiment} & \multicolumn{1}{c}{Token-level Sentiment Diffs}  \\
& \multicolumn{1}{c}{(Accuracy)} & \multicolumn{1}{c}{($F_{0.5}$)}  \\
Model Train. Data (Num. Reviews) & \textsc{Contrast} & \textsc{Contrast}  \\
\midrule
\textsc{Random} & 49.8 & 8.4 \\
\midrule
\textsc{Orig.} (3.4k) & 82.4 & 17.1 (\textsubscript{\textsc{+orig\_dev}}) \\
\textsc{Orig.+Rev.} (1.7k+1.7k) & 93.0 & 28.4 (\textsubscript{\textsc{+orig\_dev}})\\
\textsc{Orig.\textsubscript{DISJOINT}+Rev.} (1.7k+1.7k) & 91.2 & 26.9 (\textsubscript{\textsc{+orig\_dev}})\\
\midrule
\textsc{Orig.} (19k) & 81.4 & 18.0 (\textsubscript{\textsc{+orig\_dev}})\\
\textsc{Orig.+Rev.} (19k+1.7k) & 90.0 & 23.5 (\textsubscript{\textsc{+orig\_dev}})\\
\textsc{Orig.\textsubscript{DISJOINT}+Rev.} (19k-1.7k+1.7k) & 88.1 & 23.4 (\textsubscript{\textsc{+orig\_dev}})\\
\bottomrule
\end{tabular}
\end{table*}

\begin{table*}[hbt!]
\caption[Contrast Sentiment Prediction Seq. ExA]{Predicting \textit{sentiment diffs} at the token level ($F_{0.5}$) with \textsc{uniCNN+BERT} on the \textsc{Contrast} test set, applying the exemplar auditing decision rules. Predictions without accessing the support set ($\sS$) are displayed in \textcolor{red}{\texttt{red}}. Underlined results indicate $\sS$ contains additional reviews or signal not seen by the model during training. Results with access to token-level labels in $\sS$ are further highlighted in \textcolor{blue}{\underline{blue}}. $|\sS|$ is relatively small in the rightmost column. None of the models see data from the \textsc{Contrast} set dev set, either in training or in $\sS$.}
\label{table:test-results-contrast-sentiment-sequence-level-exa-rules}
\centering
%\footnotesize
\scriptsize
\begin{tabular}{l c c c c}
\toprule
\multicolumn{5}{c}{\textbf{Contrast Sets}}  \\
\midrule
& \multicolumn{4}{c}{Token-level Sentiment Diffs ($F_{0.5}$)}  \\
\cmidrule[0.75pt](lr){2-5}\\
& $\sS=$ & $\sS=$ & $\sS=$ & $\sS=$ \\
& \textsc{Orig.} & \textsc{Rev.}  & \textsc{Orig.+Rev.} & \textsc{Orig.\_dev+Rev.\_dev}\\
& (1.7k) & (1.7k) & (1.7k+1.7k) & (245+245)\\
\cmidrule[0.75pt](lr){2-2}\cmidrule[0.75pt](lr){3-3}\cmidrule[0.75pt](lr){4-4}\cmidrule[0.75pt](lr){5-5} \\
Model Train. Data (Num. Reviews) & Test: \textsc{Contrast} & Test: \textsc{Contrast} & Test: \textsc{Contrast} & Test: \textsc{Contrast}\\
\midrule
\textsc{Random} & 8.4 & 8.4 & 8.4 & 8.4\\
\midrule
\textsc{Orig.} (3.4k) & \textcolor{red}{\texttt{12.3}} & \textcolor{red}{\texttt{12.3}} & \textcolor{red}{\texttt{12.3}} & \textcolor{red}{\texttt{12.3}}\\
%\textsc{Orig.+ExA} (3.4k) & 13.7 & \underline{13.2} & \underline{13.4} & \underline{13.4}\\
\textsc{Orig.+ExAG} (3.4k) & 15.1 & \underline{18.2} & \underline{17.6} & \underline{16.4}\\
\textsc{Orig.+ExAT} (3.4k) & \textcolor{blue}{\underline{25.7}} & \textcolor{blue}{\underline{27.0}} & \textcolor{blue}{\underline{27.5}} & \textcolor{blue}{\underline{24.0}}\\
\midrule
\textsc{Orig.+Rev.} (1.7k+1.7k) & \textcolor{red}{\texttt{21.2}} & \textcolor{red}{\texttt{21.2}} & \textcolor{red}{\texttt{21.2}} & \textcolor{red}{\texttt{21.2}} \\
%\textsc{Orig.+Rev.+ExA} (1.7k+1.7k) & 26.3 & 25.5 & 26.2 & \underline{26.5} \\
\textsc{Orig.+Rev.+ExAG} (1.7k+1.7k) & 27.8 & 29.0 & 29.8 & \underline{29.1} \\
\textsc{Orig.+Rev.+ExAT} (1.7k+1.7k) & \textcolor{blue}{\underline{28.5}} & \textcolor{blue}{\underline{29.6}} & \textcolor{blue}{\underline{31.6}} & \textcolor{blue}{\underline{27.9}} \\
\midrule
\textsc{Orig.\textsubscript{DISJOINT}+Rev.} (1.7k+1.7k) & \textcolor{red}{\texttt{17.6}} & \textcolor{red}{\texttt{17.6}} & \textcolor{red}{\texttt{17.6}} & \textcolor{red}{\texttt{17.6}} \\
%\textsc{Orig.\textsubscript{DISJOINT}+Rev.+ExA} (1.7k+1.7k) & \underline{22.0} & 22.0 & \underline{21.8} & \underline{22.8} \\
\textsc{Orig.\textsubscript{DISJOINT}+Rev.+ExAG} (1.7k+1.7k) & \underline{24.7} & 25.7 & \underline{25.8} & \underline{26.0} \\
\textsc{Orig.\textsubscript{DISJOINT}+Rev.+ExAT} (1.7k+1.7k) & \textcolor{blue}{\underline{27.2}} & \textcolor{blue}{\underline{30.2}} & \textcolor{blue}{\underline{29.6}} & \textcolor{blue}{\underline{28.7}}\\
\bottomrule
\end{tabular}
\end{table*}

\subsection{Sentiment Classification \& Feature Detection: Discussion}\label{sec:sentiment-discussion}

As with error detection, on the sentiment datasets we demonstrate that we can leverage dense representation matching to update a model and to improve token-level feature detection. Remarkably, with a strong neural model and an inductive bias conducive to matching, we can start to close the distance with models trained with domain-shifted data by just updating the support set, which points to new flexibility in adapting models. However, this still requires a least some data from the distribution of the new domain to be available. When we carefully control for data distributions, robust prediction over data from unseen domain-shifted and out-of-domain distributions remains challenging, ceteris paribus, even with recently proposed data perturbation approaches, which is consistent with the broad patterns observed in the contemporaneous works of \citet{RohanEtAl-2020-NaturalDistributionShifts} and \citet{GulrajaniAndLopez-Paz2021-ERM-domain-generalization} for image data. This is a point of concern for higher-risk settings, as some amount of domain shift or subpopulation shift will invariably occur in many real-world settings.

\begin{figure}[hbt!]
    \centering
    \includegraphics[scale=0.45]{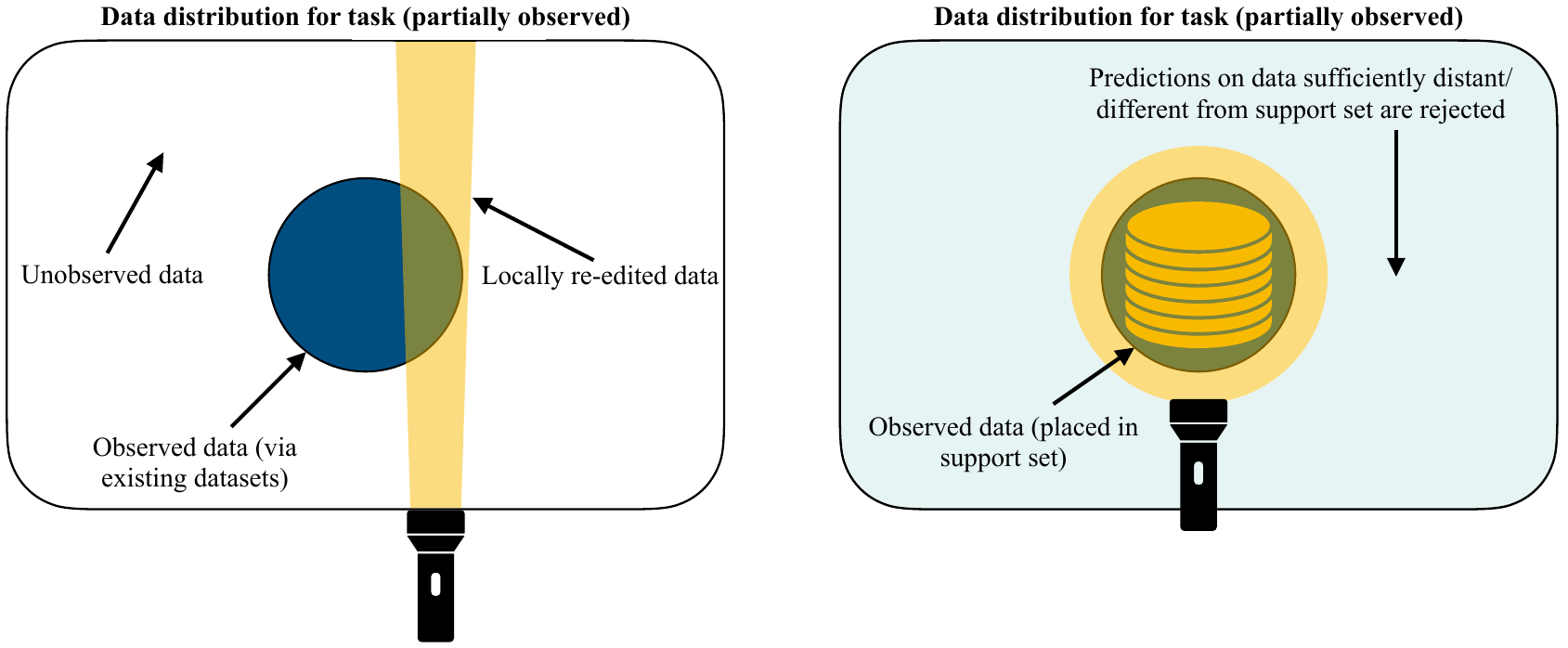}
    \caption{Local re-edits and the proposed approach for dense representation matching can be used in conjunction, but here we emphasize the contrasts for instructional purposes. Manually perturbing data around identified features, creating source-target pairs (over this small slice, illuminated by the flashlight at \textit{left}), can expand a training set and be useful for analysis; however, re-annotating in this manner can be a non-trivial task to avoid inadvertently creating annotation artifacts. As an alternative outlook for higher-risk settings (\textit{right}), we can place as much data as possible into the support set---including data not seen in training---and then conservatively only admit predictions matched closely to the support set, with flexibility over the unit of analysis using our proposed methods, sending rejected predictions to a human for further adjudication, and/or labeling.} %In this framework, confounding and spurious features can be addressed by labeling them as such in the support set, either via conditionally predicting such features (e.g., predicting a confounder, such as the domain, and introspecting the token-level features), or in high-risk settings, manually labeling at the granularity of concern.}    
    \label{fig:support-set-constraints}
\end{figure}

Faced with these challenges, we can instead constrain document-level predictions based on an interpretable token-level K-NN derived from the deep model. This combination of feature-level detection derived from document-level labels, dense matching, and heuristics that can be traced back to individual token-level predictions across the support set offers an alternative, practical approach for deploying deep models in higher-risk settings, in which we refrain from predicting over domain-shifted data and out-of-domain data over which reliable predictions and bounds remain elusive. In this way, we can refrain from predicting when necessary and then re-label, update, and as needed, re-train models in a continual loop based on these methods. For instructive purposes, we contrast such a framework with local re-edits in Figure~\ref{fig:support-set-constraints}.

%%%%%%%%%%%%%%%%%%%%%%%%%%%%%%%%%%%%%%%%%%
%%%% SECTION
%%%%%%%%%%%%%%%%%%%%%%%%%%%%%%%%%%%%%%%%%%
\section{Sentiment Data: Binary Prediction of Local Annotation Edits}\label{sec:local-annotation-detection}

In Section~\ref{sec:sentiment-experiments}, we found locally re-edited data to be useful in analyzing and evaluating feature detection for a classification task typically only labeled at the document-level. In this section, we use the same datasets to demonstrate that our proposed methods can be used to uncover subtle distributional differences across annotations, which can be used, for example, for filtering and performing quality control on datasets for training and evaluation.  

\subsection{Binary Prediction of Local Annotation Edits: Experiments}\label{sec:local-annotation-detection-experiments}
\citet{Kaushik-etal-2019-CounterfactualDataAnnotations} reports that the \textsc{BERT\textsubscript{BASE\textsubscript{uncased}}+FT} model is able to distinguish original vs. revised reviews (hereafter, ``annotator domain'') with an accuracy of about 77 percent. We investigate this further, illustrating how the proposed approach for token-level detections can be used for fine-grained text analysis.  

\subsubsection{Data: Predicting Annotator Domain (Original vs. Revised)} We assign $Y=-1$ to the original reviews and $Y=1$ to the revised reviews. We report results at the review level on varying subsets of the data, including splits by sentiment. We refer to the subset of original \textit{and} revised reviews restricted to reviews with negative sentiment with the label \textsc{(Orig.+Rev.)$\wedge$Neg.}, and similarly for other subsets. We derive token-level labels analogously to those created for sentiment in Section~\ref{sec:sentiment-experiments}, except the diffs here represent the transduction from revised reviews (source) to original reviews (target). Applicable tokens in revised reviews receive a class 1 ($y_n=1$) label, whereas tokens in original reviews are all assigned a $y_n=-1$ label. We similarly analyze the professionally annotated \textsc{Contrast} test set of \citet{Gardner-etal-2020-ContrastSets}, predicting the original reviews vs. the professionally annotated alternatives.

\subsubsection{Models} We train the \textsc{uniCNN+BERT} model on the 3414 parallel original and counterfactually-augmented revised reviews, using the 490 paired reviews of the dev set to choose the epoch with highest accuracy.

\subsection{Binary Prediction of Local Annotation Edits: Results}\label{sec:local-annotation-detection-results}

\paragraph{Predicting Annotator Domain (Original vs. Revised)}

With the \textsc{uniCNN+BERT} model, original reviews are distinguishable from counterfactually revised reviews with an accuracy of around 80\%, as shown in Table~\ref{table:test-results-counter-domain-prediction}. The revised reviews are slightly easier to distinguish in general (accuracy of 80.5 vs. 78.7). The negative reviews are particularly distinct in relative terms, with the accuracy nearly 9 points higher on the negative reviews in the combined set, with an accuracy of 84.0 vs. 75.2 for the positive reviews.  %The revised reviews and the negative reviews tend to be easier to identify as being revised, as shown by the higher accuracies for those subsets.

We further examine the particularly distinctive language used in the negative reviews using the aggregate feature extraction of Section~\ref{sec:methods-feature-extraction}. We split the dev set\footnote{The overall accuracy for annotator domain prediction is 79.4 on the dev set, which is similar to that of the test set (79.6).} according to the true document-level labels. Table~\ref{table-reviews-domain-negative-positive-total-unigrams} presents the top and lowest scoring negative class (i.e., original reviews) unigrams and positive class (i.e., revised reviews) unigrams, by total score ($\totalngram^{-}$ and $\totalngram^{+}$) for the dev set reviews for each class\footnote{The analogous $\totalngram^{-}$ and $\totalngram^{+}$ scores for \textsc{Rev.} and \textsc{Orig.}, respectively, which are not shown, exhibit patterns in the expected, corresponding directions.}, as well as the corresponding unigram frequency. We see a sharp distinction between the words most discriminative for each class. Certain unigrams, such as \texttt{not} and \texttt{bad} occur with similar frequency in the original and revised reviews, but have diametrically opposed weightings for the respective classes. It seems that words that tend to be sentiment-laden, especially those that are of negative sentiment, are particularly discriminative features for distinguishing revised reviews. In Table~\ref{table-reviews-domain-negative-positive-mean-5grams}, we show the 5-grams normalized by occurrence.\footnote{For display purposes, we have dropped subsequent n-grams with the same score, which typically just differ by a single non-discriminating word as the prefix or suffix token.} The most discriminating phrases across classes are distinct, with the contextual use of words such as ``bad'', ``not'', and ``waste'' recognized by the model as being distinctive of original vs. revised reviews.  

In Table~\ref{table-revised-reviews-highest-norm-scores} we display the top two revised reviews, ranked by $\ngram_{1:N}^{+}$, normalized by length. We have further highlighted both the ground-truth token-level domain diffs and the zero-shot sequence labeling predictions by the model (i.e., $s^{+-}_n>0$). The token-level domain diff predictions typically are subsets of the true diffs, with a focus on particularly sentiment-laden words, along the lines of what was shown in Tables~\ref{table-reviews-domain-negative-positive-total-unigrams} and \ref{table-reviews-domain-negative-positive-mean-5grams}. More generally, rather remarkably, the zero-shot sequence labeling is sufficiently effective that the approach can be used as a tool for quickly scanning through a dataset for distinctive words and phrases conditional on the document-level label, as demonstrated with additional examples in Table~\ref{table-revised-reviews-selected-awkward-phrases}. Interestingly, just reading the documents in isolation, it is not always obvious that many of the detected diffs are from revisions, yet the model is nonetheless often able to detect such subtle distributional differences.

\begin{table}[hbt!]
\caption[Counterfactually-augmented Domain Prediction]{Predicting original (\textsc{Orig.}) vs. revised (\textsc{Rev.}) data using the \textsc{uniCNN+BERT} model on the test set, with additional results subdivided by sentiment and the annotator domain classes. \textsc{Random} has an accuracy of $\approx 50$ for each subset.} %Any mistake in the \textsc{Orig.} subset is a false positive, and any mistake in the \textsc{Rev.} subset is a false negative. The revised reviews and the negative reviews tend to be easier to identify as being revised, as shown by the higher accuracies for those subsets.} 
\label{table:test-results-counter-domain-prediction}
%\centering
\footnotesize
%\scriptsize
\begin{tabular}{lcc}
\toprule
& \multicolumn{2}{c}{Review-level {\color{white}\colorbox{black}{Annotator Domain}} (\textit{Not Sentiment})}  \\
Test (Sub-)Set & Accuracy & Num. Reviews \\
\midrule
\textsc{Orig.+Rev.} & 79.6 & 976 \\
\midrule
\textsc{Orig.} & 78.7 &  488 \\
\textsc{Rev.} & 80.5 &  488 \\
\midrule
\textsc{(Orig.+Rev.)$\wedge$Neg.} & 84.0 &  488 \\
\textsc{(Orig.+Rev.)$\wedge$Pos.} & 75.2 &  488 \\
\midrule
\textsc{Orig.$\wedge$Neg.} & 84.0 & 243 \\
\textsc{Orig.$\wedge$Pos.} & 73.5 &  245 \\
\midrule
\textsc{Rev.$\wedge$Neg.} & 84.1 & 245 \\
\textsc{Rev.$\wedge$Pos.} & 77.0 &  243 \\
\bottomrule
\end{tabular}
\end{table} 

\begin{table*}[hbt!] %[b]
\caption{The top and lowest scoring negative class (i.e., original reviews) unigrams and positive class (i.e., revised reviews) unigrams, by total score ($\totalngram^{-}$ and $\totalngram^{+}$) for the dev set reviews for the respective class. We display the total score to highlight that certain unigrams, such as \texttt{not} and \texttt{bad} occur with similar frequency in the original and revised reviews, but have diametrically opposed weightings for the respective classes.}
\label{table-reviews-domain-negative-positive-total-unigrams}
\centering
\footnotesize
\begin{tabular}{lcc lcc}
\toprule
\multicolumn{6}{c}{Review-level {\color{white}\colorbox{black}{Annotator Domain}} (\textit{Not Sentiment})}  \\
\midrule
\multicolumn{3}{c}{Orig.} & \multicolumn{3}{c}{Rev.} \\
\cmidrule[0.75pt](lr){1-3}\cmidrule[0.75pt](lr){4-6}
%\midrule
unigram & $\totalngram^{-}$ score & Total Frequency & unigram & $\totalngram^{+}$ score & Total Frequency\\
\midrule
but & 41.5 & 249 & not & 61.4 & 229\\
waste & 18.5 & 22 & terrible & 54.3 & 20\\
any & 11.9 & 56 & least & 44.1 & 26\\
just & 11.0 & 112 & bad & 43.1 & 61\\
still & 8.6 & 40 & worst & 32.6 & 22\\
that & 7.7 & 394 & poor & 31.9 & 21\\
only & 7.6 & 70 & awful & 22.4 & 13\\
But & 7.6 & 42 & dislike & 20.2 & 9\\
moving & 5.8 & 7 & great & 18.5 & 69\\
completely & 5.3 & 18 & boring & 18.1 & 25\\
\midrule
\multicolumn{6}{c}{\ldots SKIPPED \ldots} \\
\midrule
hated & -1.2 & 3 & missed & -1.8 & 4\\
excited & -1.3 & 3 & without & -1.8 & 21\\
horrible & -1.4 & 5 & just & -2.1 & 97\\
worst & -1.4 & 19 & lacks & -2.2 & 3\\
usual & -1.6 & 4 & lost & -2.3 & 9\\
disliked & -1.6 & 1 & I & -2.6 & 561\\
worse & -1.9 & 8 & that & -3.1 & 395\\
hate & -1.9 & 3 & any & -4.3 & 40\\
bad & -5.5 & 64 & waste & -4.5 & 9\\
not & -7.6 & 217 & but & -10.0 & 203\\
\bottomrule
\end{tabular}
\end{table*}

\begin{table*}[hbt!] %[b]
\caption{The top and lowest scoring negative class (i.e., original reviews) 5-grams and positive class (i.e., revised reviews) 5-grams, normalized by occurrence ($\meanngram^{-}$ and $\meanngram^{+}$) for the dev set reviews for the respective class. }
\label{table-reviews-domain-negative-positive-mean-5grams}
\centering
%\footnotesize
\scriptsize
\begin{tabular}{lc lc}
\toprule
\multicolumn{4}{c}{Review-level {\color{white}\colorbox{black}{Annotator Domain}} (\textit{Not Sentiment})}  \\
\midrule
\multicolumn{2}{c}{Orig.} & \multicolumn{2}{c}{Rev.} \\
\cmidrule[0.75pt](lr){1-2}\cmidrule[0.75pt](lr){3-4}
%\midrule
5-gram & $\meanngram^{-}$ score & 5-gram & $\meanngram^{+}$ score \\
\midrule
little bit, but it still & 3.9 & his awful performance did not & 11.3\\
bit, but it still managed & 3.7 & dominated this film, his awful & 10.4\\
movie, but many elements ruined & 3.4 & Come is indeed a terrible & 10.3\\
killer down. A serious waste & 3.4 & a terrible work of speculative & 10.3 \\
this slow paced, boring waste & 3.4 & This was a very bad & 10.0\\
movie is just a waste & 3.1 & \texttt{/><br />}A terrible look at & 8.8 \\
waste of time. The most & 2.9 & dream home. \texttt{<br /><br />}A terrible & 8.8 \\
to be nice people, but & 2.8 & movie is not a lot & 8.2 \\
nice people, but can't carry & 2.8 & This movie is not a & 8.2 \\
people, but can't carry a & 2.8 & remains one of my least & 7.9\\
\midrule
\multicolumn{4}{c}{\ldots SKIPPED \ldots} \\
\midrule
film. The usual superb acting & -1.6 & either been reduced to stereo &  -1.7\\
disliked it and looking at & -1.6 & around have either been reduced & -1.7 \\
the reasons that I disliked & -1.6 & would simply be a waste & -1.7 \\
film or an even worse & -2.0 & don't waste your time and & -1.7 \\
this is such a bad & -2.5 & about lovey-dovey romance, don't waste & -1.9 \\
\bottomrule
\end{tabular}
\end{table*}

\begin{table*}[hbt!] %[b]
\caption{Top two revised reviews in the counterfactually-augmented dev set, ranked by $\ngram_{1:N}^{+}$, normalized by length. We have included the original review, \textsc{Original}, and the revised review, \textsc{True (Rev.)}, where \underline{underlines} indicate ground-truth token-level annotator domain diffs (i.e., that the token participated in a transduction between an original and revised review). We show the prediction by the \textsc{uniCNN+BERT} model to predict original vs. revised reviews, with token-level predictions \underline{underlined}, and correct predictions further highlighted in \textcolor{blue}{\underline{blue}}.}
\label{table-revised-reviews-highest-norm-scores}
\centering
%\footnotesize
\scriptsize
%\begin{tabular}{P{35mm}T{100mm}}
\begin{tabular}{P{25mm}T{100mm}}
\toprule
\multicolumn{2}{c}{Review-level {\color{white}\colorbox{black}{Annotator Domain}} (\textit{Not Sentiment})}  \\
\midrule
& Dev. Set Document 244/245\\
\textsc{Original} & \ttfamily{This is actually one of my favorite films, I would recommend that EVERYONE watches it. There is some great acting in it and it shows that not all "good" films are American....} \\
\textsc{True (Rev.)} & \ttfamily{This is actually one of my \underline{least} favorite films, I would \underline{not} recommend that \underline{ANYONE} watches it. There is some \underline{bad} acting in it and it shows that \underline{all "bad"} films are American....}\\
\textsc{uniCNN+BERT (Rev.) Len. Norm. Score: 0.164} & \ttfamily{This is actually one of my \textcolor{blue}{\underline{least}} favorite films, I would \textcolor{blue}{\underline{not}} recommend that ANYONE watches it. There is some bad acting in it and it shows that all "bad" films are American....}\\
\midrule
& Dev. Set Document 266/267\\
\textsc{Original} & \ttfamily{One of the great classic comedies. Not a slapstick comedy, not a heavy drama. A fun, satirical film, a buyers beware guide to a new home. /> />Filled with great characters all of whom, Cary Grant is convinced, are out to fleece him in the building of a dream home. /> />A great look at life in the late 40's. /> />} \\
\textsc{True (Rev.)} & \ttfamily{One of the \underline{bad} classic comedies. Not a slapstick comedy, not a heavy drama. A \underline{boring, unfunny} film, a buyers beware guide to a new home. /> />Filled with \underline{terrible} characters all of whom, Cary Grant is \underline{falsely} convinced, are out to fleece him in the building of a dream home. /> />A \underline{terrible} look at life in the late 40's. /> />}\\
\textsc{uniCNN+BERT (Rev.) Len. Norm. Score: 0.133} & \ttfamily{One of the bad classic comedies. Not a slapstick comedy, not a heavy drama. A \textcolor{blue}{\underline{boring,}} unfunny film, a buyers beware guide to a new home. /> />Filled with terrible characters all of whom, Cary Grant is falsely convinced, are out to fleece him in the building of a dream home. /> />A \textcolor{blue}{\underline{terrible}} look at life in the late 40's. /> />}\\
\bottomrule
\end{tabular}
\end{table*}

\paragraph{Predicting Annotator Domain (Original vs. Professional Revisions)}

The model is nearly as effective at distinguishing the professionally annotated reviews as the crowd-sourced revised reviews, with the overall accuracy only a couple of points lower, as shown in Table~\ref{table:test-results-contrast-sets-domain-prediction}, even though the model only sees crowd-sourced revisions in training and development. The negative reviews are again easier to distinguish overall, but in this case we see that this is driven by accuracy on the original reviews, which are more readily distinguished. This might be attributable to the effects of the domain shift, with the original reviews being seen in training, while the professionally annotated counterparts are not. As with the counterfactually-augmented edits, without such model assistance, it is often not obvious that a review has been revised, especially given the otherwise informal language of movie reviews. However, the class-conditional feature detection is strong enough that the token-level predictions can be visualized and some of the discriminative words and phrases participating in the diffs identified, as shown in Table~\ref{table-contrast-sets-reviews-selected-awkward-phrases}.

\subsection{Prediction of Local Annotation Edits: Discussion}

With effective zero-shot sequence labeling, we gain a straightforward means of aggregating features from a deep network when only given document-level labels. As we have shown, this can be used to analyze text datasets, detecting rather subtle distributional differences that are not readily perceptible without such model assistance, at least at scale. Deep networks are typically viewed as strong predictors at the unit of analysis of the training set's labels; with the mechanism proposed here, we gain a means of leveraging that discriminative ability at lower resolutions to analyze the input data.

\begin{table}[hbt!]
\caption[Contrast Sets Domain Prediction]{Predicting original (\textsc{Orig.}) vs. revised contrast set (\textsc{Contrast}) data using the \textsc{uniCNN+BERT} model on the test set, with additional results subdivided by sentiment and the annotator domain classes. \textsc{Random} has an accuracy of $\approx 50$ for each subset.} 
\label{table:test-results-contrast-sets-domain-prediction}
%\centering
\footnotesize
%\scriptsize
\begin{tabular}{lcc}
\toprule
\multicolumn{3}{c}{\textbf{Contrast Sets}}  \\
\midrule
& \multicolumn{2}{c}{Review-level {\color{white}\colorbox{black}{Annotator Domain}} (\textit{Not Sentiment})}  \\
Test (Sub-)Set & Accuracy & Num. Reviews \\
\midrule
\textsc{Orig.+Contrast} & 77.8 & 976 \\
\midrule
\textsc{Orig.} & 78.7 &  488 \\
\textsc{Contrast} & 76.8 &  488 \\
\midrule
\textsc{(Orig.+Contrast)$\wedge$Neg.} & 78.7 &  488 \\
\textsc{(Orig.+Contrast)$\wedge$Pos.} & 76.8 &  488 \\
\midrule
\textsc{Orig.$\wedge$Neg.} & 84.0 & 243 \\
\textsc{Orig.$\wedge$Pos.} & 73.5 &  245 \\
\midrule
\textsc{Contrast$\wedge$Neg.} & 73.5 & 245 \\
\textsc{Contrast$\wedge$Pos.} & 80.2 &  243 \\
\bottomrule
\end{tabular}
\end{table}

%%%%%%%%%%%%%%%%%%%%%%%%%%%%%%%%%%%%%%%%%%
%%%% SECTION
%%%%%%%%%%%%%%%%%%%%%%%%%%%%%%%%%%%%%%%%%%
\section{Discussion}

This new facility for dense representation matching at resolutions of the input more fine-grained than available labels is a substantive departure from existing approaches in computational linguistics, providing new flexibility for locally updating a model and analyzing datasets under the model. It draws a connection between attention-style mechanisms and the older distance metric learning literature \cite[inter alia]{WeinbergerAndSaul-LMNN-2009}, relying on the inductive bias of the CNN to learn summarized representations of the expressive deep network for subsequent matching via simple Euclidean distances. Fortunately from an efficient compute perspective, this works well when training with standard cross-entropy losses against the available labels without resorting to expensive supervised contrastive losses searching through representations during initial training. When a stronger sense of interpretability is needed, we can then subsequently train an effective K-NN approximation with just 3 learnable parameters from the frozen representations. 

Prototypical networks \cite{SnellEtAl-NIPS2017-PrototypicalNets} and matching networks \cite{VinyalsN-EtAl-NIPS-2016-MatchingNetworks} can also be updated by modifying a support set, but the means of doing so are markedly different than we have proposed, motivated by different intended use cases. Critical for NLP settings, we are concerned with fine-grained feature detection, which necessitates the proposed indirect approach for deriving predictions and representations from an imputation-trained deep network, and a different approach for training. Additionally, unlike prototypical networks, we perform matching against every instance (in fact, every \textit{token}) in the support set, rather than class means, which is a strength rather than a weakness for the intended interpretability and dataset analysis applications. Finally, matching networks can also be viewed as a particular weighted K-NN. In contrast, our K-NN approximation of an already trained model is proposed as a parsimonious, interpretable replacement of the original model, and is trained accordingly.\footnote{With regard to model approximations, there is also an indirect connection to work relating kernel machines to neural architectures and vice-versa \cite[inter alia]{NIPS2009-ArcCosKernels,NIPS-2017-ArcCosEmpirical}.} %which we leave for future work.}

%%%%%%%%%%%%%%%%%%%%%%%%%%%%%%%%%%%%%%%%%%
%%%% SECTION
%%%%%%%%%%%%%%%%%%%%%%%%%%%%%%%%%%%%%%%%%%
\section{Conclusion}

Deep networks are typically viewed as strong predictors that are otherwise immutable and inscrutable black boxes, with the non-identifiable parameters running into the millions and higher. In this context, we have demonstrated a series of approaches toward a more actionable understanding of a deep network over its input data. We have shown a kernel-width-one CNN and a linear layer over a deep network is effective for deriving token-level predictions when only given document-level labels for training. This approach for class-conditional feature detection enables dense representation matching against a support set with known labels, which can be used with inference-time decision rules to constrain predictions. Additionally, we have shown that we can altogether replace a model's output with an interpretable weighting over instances with known labels without loss of predictive effectiveness. In this way, we gain sequence labeling at varying label resolutions; local updatability of a model without re-training; interpretable token-level constraints over domain-shifted and out-of-domain data; and more generally, a model-assisted means for uncovering patterns in large datasets that may not be readily detectable at scale without the expressive, deep networks. 

\begin{acknowledgments}
We thank the reviewers for their feedback and suggestions.
\end{acknowledgments}

%%%%%%%%%%%%%%%%%%%%%%%%%%%%%%%%%%%%%%%%%%%%%%%%%%%%%%
%%%%%%%%%%%%%%%%%%%%%%%%%%%%%%%%%%%%%%%%%%%%%%%%%%%%%%
\clearpage
\appendix
%%%%%%%%%%%%%%%%%%%%%%%%%%%%%%%%%%%%%%%%%%%%%%%%%%%%%%
%%%%%%%%%%%%%%%%%%%%%%%%%%%%%%%%%%%%%%%%%%%%%%%%%%%%%%

%%%%%%%%%%%%%%%%%%%%%%%%%%%%%%%%%%%%%%%%%%
%%%% APPENDIXSECTION
%%%%%%%%%%%%%%%%%%%%%%%%%%%%%%%%%%%%%%%%%%
\appendixsection{Contents}

In Appendix B, C, and D, we provide additional results and output for the experiments on the grammatical error detection task, the sentiment datasets, and for the experiments predicting annotator re-edits, respectively.

%%%%%%%%%%%%%%%%%%%%%%%%%%%%%%%%%%%%%%%%%%
%%%% APPENDIXSECTION
%%%%%%%%%%%%%%%%%%%%%%%%%%%%%%%%%%%%%%%%%%
\appendixsection{Grammatical Error Detection Analysis and Examples}

Table~\ref{appendix1-table-FCE-test-examples-part1} shows five random examples of original sentences from the FCE test set and the corresponding labeled outputs from the \textsc{cnn}, \textsc{uniCNN+BERT}, \textsc{uniCNN+BERT+mm}, and \textsc{uniCNN+BERT+S*} models.

%\paragraph{Exemplar Auditing}
%\ref{appendix1-table-FCE-test-exemplar-audit-part1-uniCNNBERT} \textsc{uniCNN+BERT}, 
Tables~\ref{appendix1-table-FCE-test-exemplar-audit-part1-uniCNNBERTmm} and~\ref{appendix1-table-FCE-test-exemplar-audit-part1-uniCNNBERTsupervised} show the nearest matches used for the proposed inference-time decision rules for the first three sentences with ground-truth grammatical errors from Table~\ref{appendix1-table-FCE-test-examples-part1} for the \textsc{uniCNN+BERT+mm} and \textsc{uniCNN+BERT+S*} models, respectively. We have provided the exemplar tokens and associated sentences from the support set (here, consisting of the FCE training set) wherever the model makes a positive prediction. For reference, we have also provided the sentence corresponding to the exemplar representation for any tokens marked in the ground-truth labels but missed by the model. The qualitative analysis is consistent with the quantitative results in the main text: When the test prediction is in the same direction of the prediction of the exemplar from the support set, the corresponding contexts, and the exemplar word itself---which is not always a verbatim lexical match---are often similar, particularly when the $L^2$ distances are low.

Table~\ref{appendix1-table-FCE-training-positive-ngrams} contains the unigram positive class n-grams normalized by occurrence ($\meanngram^{+}$) for the training sentences for which $Y=1$. The top scoring such unigrams constitute a relatively sharp list of misspellings. We also include the lowest scoring such unigrams at the bottom of the table, as a check on our featuring scoring method. The ranked features are as we would expect, with the lowest scoring unigrams being names and other words that are otherwise correctly spelled.

Table~\ref{table:fce-news2k-domain-shifted-test-magnitude-quantiles} compares the K-NN output with that of the original model, \textsc{uniCNN+BERT+mm}, on the domain-shifted test set, as with Figure~\ref{fig:fcenews2k-test-magnitude-comparator} in the main text. 

\clearpage
%\resizebox{0.95\textwidth}{!}{ % If your table exceeds the column or page width, use this command to reduce it slightly

\begin{table*}[!htbp] %[b]
\caption{Five random sentences from the FCE test set. The ground-truth labeled sentences are marked \textsc{True}, with ground-truth token-level labels underlined. In the case of model output, underlines indicate predicted error labels. Note that sentence 1551, as with the other sentences, is verbatim from the gold test set.}
\label{appendix1-table-FCE-test-examples-part1}
\centering
\footnotesize
\begin{tabular}{P{35mm}T{95mm}}
\toprule
 & Sentence 174\\
\textsc{True} & \ttfamily{There \underline{are} some \underline{informations} you have asked me about .}\\
\textsc{cnn}  & \ttfamily{There \underline{are some informations you have} asked me about .}\\
\textsc{uniCNN+BERT}  & \ttfamily{There are some \underline{informations} you have asked me about .}\\
\textsc{uniCNN+BERT+mm}  & \ttfamily{There are some \underline{informations} you have asked me about .}\\
\textsc{uniCNN+BERT+S*}  & \ttfamily{There are some \underline{informations} you have asked me about .}\\
\midrule
& Sentence 223\\
\textsc{True} & \ttfamily{There is space for about five hundred people .}\\
\textsc{cnn}  &  \ttfamily{There is space for \underline{about five hundred people .}}\\
\textsc{uniCNN+BERT}  & \ttfamily{There is space for about five hundred people .}\\
\textsc{uniCNN+BERT+mm}  & \ttfamily{There is space for about five hundred people .}\\
\textsc{uniCNN+BERT+S*}  & \ttfamily{There is space for about five hundred people .}\\
\midrule
& Sentence 250\\
\textsc{True} & \ttfamily{It is n't easy \underline{giving} an answer \underline{at} this question .}\\
\textsc{cnn}  &  \ttfamily{It \underline{is} n't easy \underline{giving an answer} at this question .}\\
\textsc{uniCNN+BERT}  & \ttfamily{It is n't easy giving an \underline{answer at} this \underline{question} .}\\
\textsc{uniCNN+BERT+mm}  & \ttfamily{It is n't easy giving an answer \underline{at} this question .}\\
\textsc{uniCNN+BERT+S*}  & \ttfamily{It is n't easy giving an answer \underline{at} this question .}\\
\midrule
& Sentence 1302\\
\textsc{True} & \ttfamily{Your group has been booked \underline{in Palace} Hotel which is one of the most comfortable hotels in London .}\\
\textsc{cnn}  &  \ttfamily{\underline{Your group has been booked in Palace Hotel} which is one of \underline{the most comfortable} hotels in London .}\\
\textsc{uniCNN+BERT}  & \ttfamily{Your group has been booked \underline{in} Palace Hotel \underline{which} is one of the most \underline{comfortable} hotels in London .}\\
\textsc{uniCNN+BERT+mm}  & \ttfamily{Your group has been booked \underline{in} Palace Hotel which is one of the most comfortable hotels in London .}\\
\textsc{uniCNN+BERT+S*}  & \ttfamily{Your group has been booked in \underline{Palace} Hotel which is one of the most comfortable hotels in London .}\\
\midrule
& Sentence 1551\\
\textsc{True} & \ttfamily{By the way \underline{you} can visit the}\\
\textsc{cnn}  &  \ttfamily{\underline{By the way} you \underline{can visit the}}\\
\textsc{uniCNN+BERT}  & \ttfamily{By the way you can visit the}\\
\textsc{uniCNN+BERT+mm}  & \ttfamily{By the way you can visit the}\\
\textsc{uniCNN+BERT+S*}  & \ttfamily{By the way you can visit the}\\
\bottomrule
\end{tabular}
\end{table*}

\begin{table*} %[b]
\caption{Exemplar auditing output for three sentences from Table~\ref{appendix1-table-FCE-test-examples-part1} for the \textsc{uniCNN+BERT+mm} model. Ground-truth labeled sentences are marked \textsc{True} with ground-truth token-level labels underlined. Underlines in the  \textsc{uniCNN+BERT+mm} rows indicate predictions. We show the exemplars for the predicted tokens and for reference, any true token labels missed by the model. In both cases, the exemplar tokens from training are labeled by the index into the test sentence, as indicated in brackets. The Euclidean distance between the test token and the exemplar token is labeled with \textsc{Exemplar Dist.} The full training sentence for the exemplar is provided, with underlines indicating ground-truth labels in the case of \textsc{Exemplar True} and training predications from \textsc{uniCNN+BERT+mm} in the case of \textsc{Exemplar Pred.}}
\label{appendix1-table-FCE-test-exemplar-audit-part1-uniCNNBERTmm}
\centering
\footnotesize
\begin{tabular}{P{35mm}T{95mm}}
\toprule
 & Sentence 174\\
\textsc{True} & \ttfamily{There \underline{are}[1] some \underline{informations}[3] you have asked me about .}\\
\textsc{uniCNN+BERT+mm}  & \ttfamily{There are some \underline{informations}[3] you have asked me about .}\\
\textsc{Exemplar [1] Dist.}  & 41.1\\
\textsc{Exemplar [1] True}  & \ttfamily{But , there are[1] three things which I would like to tell you .}\\
\textsc{Exemplar [1] Pred.}  & \ttfamily{But , there are[1] three things which I would like to tell you .}\\
\textsc{Exemplar [3] Dist.}  & 34.0\\
\textsc{Exemplar [3] True}  & \ttfamily{I am very glad to hear that and would like to tell you all the \underline{informations}[3] you need to know from me .}\\
\textsc{Exemplar [3] Pred.}  & \ttfamily{I am very glad to hear that and would like to tell you all the \underline{informations}[3] you need to know from me .}\\
\midrule
& Sentence 250\\
\textsc{True} & \ttfamily{It is n't easy \underline{giving}[4] an answer \underline{at}[7] this question .}\\
\textsc{uniCNN+BERT+mm}  & \ttfamily{It is n't easy giving an answer \underline{at}[7] this question .}\\
\textsc{Exemplar [4] Dist.}  & 71.8 \\
\textsc{Exemplar [4] True}  & \ttfamily{It 's very difficult describing[4] all \underline{emotion} I felt .}\\
\textsc{Exemplar [4] Pred.}  & \ttfamily{It 's very difficult describing[4] \underline{all} \underline{emotion} I felt .}\\
\textsc{Exemplar [7] Dist.}  & 63.3\\
\textsc{Exemplar [7] True}  & \ttfamily{I 'm going to reply \underline{at}[7] your question .}\\
\textsc{Exemplar [7] Pred.}  & \ttfamily{I 'm going to reply \underline{at}[7] your question .}\\
%\textsc{Exemplar [1] Dist.}  & \\
%\textsc{Exemplar [1] True}  & \ttfamily{}\\
%\textsc{Exemplar [1] Pred.}  & \ttfamily{}\\
\midrule
& Sentence 1302\\
\textsc{True} & \ttfamily{Your group has been booked \underline{in}[5] \underline{Palace}[6] Hotel which is one of the most comfortable hotels in London .}\\
\textsc{uniCNN+BERT+mm}  &  \ttfamily{Your group has been booked \underline{in}[5] Palace Hotel which is one of the most comfortable hotels in London .}\\
\textsc{Exemplar [5] Dist.}  & 57.0 \\
\textsc{Exemplar [5] True}  & \ttfamily{Secondly I would prefer to \underline{be} \underline{accommodate} in[5] \underline{log} \underline{cabins} .}\\
\textsc{Exemplar [5] Pred.}  & \ttfamily{Secondly I would prefer to be \underline{accommodate} in[5] log cabins .}\\
\textsc{Exemplar [6] Dist.}  & 59.6\\
\textsc{Exemplar [6] True}  & \ttfamily{I insisted on going to your theatre , to the Circle[6] Theatre , because I have heard that it is one of the best theatres in London .}\\
\textsc{Exemplar [6] Pred.}  & \ttfamily{I insisted on going to your theatre , to the Circle[6] Theatre , because I have heard that it is one of the best theatres in London .}\\
\bottomrule
\end{tabular}
\end{table*}
%%%%%%%%%%%%%%%%%% Exemplar auditing tables -- \textsc{uniCNN+BERT+mm}

%%%%%%%%%%%%%%%%%% Exemplar auditing tables -- \textsc{uniCNN+BERT+S*}
\begin{table*} %[b]
\caption{Exemplar auditing output for three sentences from Table~\ref{appendix1-table-FCE-test-examples-part1} for the \textsc{uniCNN+BERT+S*} model. Ground-truth labeled sentences are marked \textsc{True} with ground-truth token-level labels underlined. Underlines in the \textsc{uniCNN+BERT+S*} rows indicate predictions. We show the exemplars for the predicted tokens and for reference, any true token labels missed by the model. In both cases, the exemplar tokens from training are labeled by the index into the test sentence, as indicated in brackets. The Euclidean distance between the test token and the exemplar token is labeled with \textsc{Exemplar Dist.} The full training sentence for the exemplar is provided, with underlines indicating ground truth labels in the case of \textsc{Exemplar True} and training predications from \textsc{uniCNN+BERT+S*} in the case of \textsc{Exemplar Pred.}}
\label{appendix1-table-FCE-test-exemplar-audit-part1-uniCNNBERTsupervised}
\centering
\footnotesize
\begin{tabular}{P{35mm}T{95mm}}
\toprule
 & Sentence 174\\
\textsc{True} & \ttfamily{There \underline{are}[1] some \underline{informations}[3] you have asked me about .}\\
\textsc{uniCNN+BERT+S*}  & \ttfamily{There are some \underline{informations}[3] you have asked me about .}\\
\textsc{Exemplar [1] Dist.}  & 32.7\\
\textsc{Exemplar [1] True}  & \ttfamily{But , there are[1] three things which I would like to tell you .}\\
\textsc{Exemplar [1] Pred.}  & \ttfamily{But , there are[1] three things which I would like to tell you .}\\
\textsc{Exemplar [3] Dist.}  & 24.0\\
\textsc{Exemplar [3] True}  & \ttfamily{I am very glad to hear that and would like to tell you all the \underline{informations}[3] you need to know from me .}\\
\textsc{Exemplar [3] Pred.}  & \ttfamily{I am very glad to hear that and would like to tell you all the \underline{informations}[3] you need to know from me .}\\
\midrule
& Sentence 250\\
\textsc{True} & \ttfamily{It is n't easy \underline{giving}[4] an answer \underline{at}[7] this question .}\\
\textsc{uniCNN+BERT+S*}  & \ttfamily{It is n't easy giving an answer \underline{at}[7] this question .}\\
\textsc{Exemplar [4] Dist.}  & 54.6 \\
\textsc{Exemplar [4] True}  & \ttfamily{To say nothing about his or her giving[4] advice !}\\
\textsc{Exemplar [4] Pred.}  & \ttfamily{To say nothing about his or her giving[4] advice !}\\
\textsc{Exemplar [7] Dist.}  & 44.0\\
\textsc{Exemplar [7] True}  & \ttfamily{I 'm going to reply \underline{at}[7] your question .}\\
\textsc{Exemplar [7] Pred.}  & \ttfamily{I 'm going to reply \underline{at}[7] your question .}\\
%\textsc{Exemplar [1] Dist.}  & \\
%\textsc{Exemplar [1] True}  & \ttfamily{}\\
%\textsc{Exemplar [1] Pred.}  & \ttfamily{}\\
\midrule
& Sentence 1302\\
\textsc{True} & \ttfamily{Your group has been booked \underline{in}[5] \underline{Palace}[6] Hotel which is one of the most comfortable hotels in London .}\\
\textsc{uniCNN+BERT+S*}  &  \ttfamily{Your group has been booked in \underline{Palace}[6] Hotel which is one of the most comfortable hotels in London .}\\
\textsc{Exemplar [5] Dist.}  & 43.6 \\
\textsc{Exemplar [5] True}  & \ttfamily{\underline{Accommodation} \underline{in}[5] log cabins would be better for me , because they are more comfortable .}\\
\textsc{Exemplar [5] Pred.}  & \ttfamily{\underline{Accommodation} in[5] log cabins would be better for me , because they are more comfortable .}\\
\textsc{Exemplar [6] Dist.}  & 46.8\\
\textsc{Exemplar [6] True}  & \ttfamily{There will be The London Fashion and Leisure Show in Central[6] Exhibition Hall on the 14th of March .}\\
\textsc{Exemplar [6] Pred.}  & \ttfamily{There will be The London Fashion and Leisure Show in Central[6] Exhibition Hall on the 14th of March .}\\
\bottomrule
\end{tabular}
\end{table*}
%%%%%%%%%%%%%%%%%% Exemplar auditing tables -- \textsc{uniCNN+BERT+S*}

%%%%%%%%%%%%%%%%%%%%%%%%%%%%%%%%%%%%%%%%%%%%%%%%%%%%%%
%%%%%%%%%%%%%%%%%% Exemplar auditing tables - END

\begin{table*} %[b]
\caption{The top and lowest scoring unigram positive class n-grams normalized by occurrence ($\meanngram^{+}$) for the training sentences that are marked as incorrect (i.e., belonging to the positive class) for the \textsc{uniCNN+BERT} model.}
\label{appendix1-table-FCE-training-positive-ngrams}
%\centering
\footnotesize
\begin{tabular}{lcc}
\toprule
%\midrule
unigram & $\meanngram^{+}$ score & Total Frequency\\
\midrule
wating &  22.5 & 1\\
noize &  21.9 & 1\\
exitation &  21.5 & 1\\
exitement &  21.2 & 1\\
toe &  20.1 & 1\\
fite &  20.0 &1\\
ofer &  20.0 & 2\\
n &  19.7 & 5\\
intents &  18.6 & 1\\
wit &  17.7 & 2\\
defences &  17.5 & 1\\
meannes &  17.5 & 1\\
baying &  17.3 & 1\\
saing &  17.1 & 2\\
dipends &  17.0 & 1\\
lair &  16.7 & 2\\
torne &  16.7 & 1\\
farther &  16.2 & 1\\
andy &  16.0 & 1\\
seasonaly &  15.9 & 1\\
remainds &  15.6 & 1\\
sould &  15.5 & 4\\
availble &  15.5 & 3\\
\midrule
&\ldots SKIPPED \ldots&\\
\midrule
sixteen & -1.7 & 3\\
Uruguay & -1.7 & 1\\
Jose & -1.7 & 1\\
leg & -1.7 & 3\\
Joseph & -2.0 & 1\\
deny & -2.1 & 1\\
Sandre & -2.2 & 1\\
leather & -2.4 & 2\\
shoulder & -2.6 & 1\\
apartheid & -2.8 & 1\\
tablets & -2.8 & 1\\
Martial & -3.0 & 1\\
Lorca & -3.1 & 1\\
\bottomrule
\end{tabular}
\end{table*}

\begin{table}
\caption[Model and KNN magnitude quantiles]{The original model (\textsc{uniCNN+BERT+mm}) output, $f(\cdot)$, and the K-NN approximation output, $f(\cdot)^{KNN}$, as comparative measures of prediction reliability on the domain-shifted \textsc{FCE+news2k} test set. The K-NN only has access to the original FCE training set. Quantiles are constructed by equally dividing the data after sorting based on the magnitude of the output, separated by class. When considering all of the data (4th quartile), the K-NN is already a modestly stronger predictor, but the difference amplifies with the smaller subsets because the K-NN output is a slightly stronger measure of prediction uncertainty and/or a stronger predictor conditioned on output magnitude, with relatively more of the correct predictions clustered at higher magnitudes. The K-NNs of the remaining models also track prediction reliability at least as closely as that of the original models in similar oracle sorting, as shown in Figure~\ref{fig:fcenews2k-test-magnitude-comparator}, with the advantage that the K-NNs' model terms are readily inspectable and interpretable, as described in the main text.} 
\label{table:fce-news2k-domain-shifted-test-magnitude-quantiles}
%\centering
%\footnotesize
\scriptsize
%\small
\begin{tabular}{lcccc c c c}
\toprule
& \multicolumn{4}{c}{Quantiles} & & \\
\cmidrule[0.75pt](lr){2-5} \\
Model & 1st & 2nd & 3rd & 4th & Metric & Threshold & Prediction \\
& \multicolumn{4}{c}{$\rightarrow$ \textit{cumulative} $\rightarrow$} & & \\
\midrule
\multicolumn{8}{l}{ \textsc{uniCNN+BERT+mm+K$_8$NN\textsubscript{dist.}}} \\
\cmidrule[0.5pt](lr){2-8} \\
 %& 0.91 & 0.52 & 0.24 &  0.00  &   & $\left | f(\cdot)^{KNN}_{test} \right |$ & $\hat y^{KNN}_{test}=1$  \\
 &  &  &  &   &   &   & $\hat y^{KNN}=1$ \\
  &  \lightgraybox{0.91} & \lightgraybox{0.52} &  \lightgraybox{0.24} &   \lightgraybox{0.00}  &   &  \lightgraybox{$> \left | f(\cdot)^{KNN} \right |$} &   \\
 & 56.3 & 44.2 & 36.7 & 32.1   & $F_{0.5}$  &  &    \\
& 50.7 & 38.8 & 31.7 & 27.4  &  Acc. &   &  \\
& 1459 & 2918 & 4377 & 5837    &  $N$ &   &  \\
\cmidrule[0.5pt](lr){2-8} \\
 &  &  &  &   &   &   & $\hat y^{KNN}=0$ \\
& \lightgraybox{1.32} & \lightgraybox{1.11} & \lightgraybox{0.86} &  \lightgraybox{0.00}  &   &  \lightgraybox{$>\left | f(\cdot)^{KNN} \right |$} &   \\
& 97.8 & 97.1 & 96.0 & 94.6  &  Acc. &   &  \\
& 21690 & 43380 & 65070 &  86760   &  $N$ &   &  \\
\midrule
\multicolumn{8}{l}{ \textsc{uniCNN+BERT+mm}} \\
\cmidrule[0.5pt](lr){2-8} \\
&  &  &  &   &   &   & $\hat y=1$ \\
  &  \lightgraybox{16.55} & \lightgraybox{7.34} &  \lightgraybox{2.80} &   \lightgraybox{0.00}  &   &  \lightgraybox{$>\left | f(\cdot) \right |$} &   \\
 & 48.7 & 38.7 & 32.8 & 29.5    & $F_{0.5}$  &  &    \\
& 43.1 & 33.5 & 28.1 & 25.0   &  Acc. &   &  \\
& 1831 & 3662 & 5493 &   7327  &  $N$ &   &  \\
\cmidrule[0.5pt](lr){2-8} \\
&  &  &  &   &   &   & $\hat y=0$ \\
  &  \lightgraybox{12.13} & \lightgraybox{12.10} &  \lightgraybox{9.25} &   \lightgraybox{0.00}  &   &  \lightgraybox{$>\left | f(\cdot) \right |$} &   \\
& 96.4 & 96.2 & 95.7 &  94.8 &  Acc. &   &  \\
& 21317 & 42634 & 63951 &  85270   &  $N$ &   &  \\
\bottomrule
\end{tabular}
\end{table} 

%%%%%%%%%%%%%%%%%%%%%%%%%%%%%%%%%%%%%%%%%%
%%%% APPENDIXSECTION
%%%%%%%%%%%%%%%%%%%%%%%%%%%%%%%%%%%%%%%%%%
\clearpage
\appendixsection{Sentiment Analysis}

\paragraph{Sentiment Diffs for Token-Level Detection} An example of the process to create the token-level detection labels for the sentiment datasets is shown in Table~\ref{table-sentiment-diffs-data-creation}. Note that the in-line diffs of the first row are used for data creation, but are not subsequently directly used in training or inference. The diffs are guaranteed to transduce to the source and target and the resulting positive class labels often correspond to positive sentiment. Occasionally there are edge cases created by the diff process and/or the underlying data for which an independent annotator tasked with labeling positive words might conceivably label differently. For example, in this review, ``not`` is assigned to the positive class, which is consistent with the original and revised diff of the reviews. 

%\clearpage

\begin{table*} %[b]
\caption{Example of creating the ground-truth token-level sentiment features diffs data from parallel source (positive sentiment, $Y=1$) and target (negative sentiment, $Y=-1$) data. Source-target diffs that transduce to $Y=1$ are colored blue, and those that transduce to $Y=-1$ are colored red. Tokens with positive class token-level feature labels ($y_n=1$) are \underline{underlined} in the second row. Under this convention, the corresponding negative review (the final row) is never assigned positive token labels (i.e., the colored red tokens and all other non-blue tokens are assigned $y_n=-1$).}
\label{table-sentiment-diffs-data-creation}
\centering
%\footnotesize
\scriptsize
\begin{tabular}{P{35mm}T{95mm}}
\toprule
\multicolumn{2}{c}{\textbf{Sentiment Features Diffs Data Creation (Ground-Truth Labels)}}  \\
\midrule
%& Example 1\\
source (positive sentiment) to target (negative sentiment) & \ttfamily{I saw this in the summer of 1990. I'm still \textcolor{blue}{<del> amazed </del>} \textcolor{red}{<ins> annoyed </ins>} by how \textcolor{blue}{<del> good </del>} \textcolor{red}{<ins> bad </ins>} this movie is in 2001.<br /><br \textcolor{blue}{<del> />Incredible </del>} \textcolor{red}{<ins> />Implausible </ins>} plot. You'd have to be a child to think this could \textcolor{blue}{<del> not </del>} happen.<br /><br />I'm just really \textcolor{blue}{<del> amazed </del>} \textcolor{red}{<ins> annoyed </ins>} by it. \textcolor{blue}{<del> Definitely </del>} \textcolor{red}{<ins> Don't </ins>} see this.} \\
\midrule
Resulting ground-truth labels (positive review: $Y=1$) & \ttfamily{I saw this in the summer of 1990. I'm still \textcolor{blue}{\underline{amazed}} by how \textcolor{blue}{\underline{good}} this movie is in 2001.<br /><br \textcolor{blue}{\underline{ />Incredible}} plot. You'd have to be a child to think this could \textcolor{blue}{\underline{not}} happen.<br /><br />I'm just really \textcolor{blue}{\underline{amazed}} by it. \textcolor{blue}{\underline{Definitely}} see this.} \\
\midrule
Resulting ground-truth labels (negative review: $Y=-1$) & \ttfamily{I saw this in the summer of 1990. I'm still \textcolor{red}{annoyed} by how \textcolor{red}{bad} this movie is in 2001.<br /><br \textcolor{red}{/>Implausible} plot. You'd have to be a child to think this could happen.<br /><br />I'm just really \textcolor{red}{annoyed} by it. \textcolor{red}{Don't} see this.} \\
\bottomrule
\end{tabular}
\end{table*}

\begin{table*}[hbt!] %[t]
\caption[Counterfactually-augmented Sentiment Prediction BERT Comparisons]{Accuracy results for predicting \textit{sentiment} on the original (\textsc{Orig.}) and revised (\textsc{Rev.}) test sets. These are reference results placing the proposed models in the context of fine-tuning the Transformer parameters. These models are all trained on the full original training set (19k) \textit{and} the revised training set (1.7k). The results for \textsc{BERT\textsubscript{BASE\textsubscript{uncased}}+FT}, which fine-tunes the \textsc{BERT\textsubscript{BASE\textsubscript{uncased}}} parameters, are those of \citet{Kaushik-etal-2019-CounterfactualDataAnnotations}.} 
\label{table:test-results-sentiment-review-level-bert-ft-comparison}
%\centering
\footnotesize
%\scriptsize
\begin{tabular}{lcc}
\toprule
& \multicolumn{2}{c}{Review-level Sentiment (Accuracy)} \\
Model & \textsc{Orig.} & \textsc{Rev.} \\
\midrule
 \textsc{BERT\textsubscript{BASE\textsubscript{uncased}}+FT}  & 93.2 & 93.9 \\
 \textsc{uniCNN+BERT\textsubscript{BASE\textsubscript{uncased}}} & 91.8 & 91.4 \\
 \textsc{uniCNN+BERT\textsubscript{BASE}} & 92.2 & 93.4 \\
\textsc{uniCNN+BERT} & 93.0 & 94.3 \\
\bottomrule
\end{tabular}
\end{table*}

\begin{table*}[hbt!]
\caption[Counterfactually-augmented Sentiment Prediction Seq. Out-of-domain SemEval 2017]{Predicting \textit{sentiment} on out-of-domain data, the SemEval-2017 Task 4a test set, with 
\textsc{uniCNN+BERT} and \textsc{uniCNN+BERT\textsubscript{BASE\textsubscript{uncased}}} (\textsubscript{BASE\textsubscript{uncased}}).} 
\label{table:test-results-sentiment-semeval2017}
%\centering
\footnotesize
%\scriptsize
\begin{tabular}{lc}
\toprule
& \multicolumn{1}{c}{Review-level Sentiment (Accuracy)}\\
%& \multicolumn{1}{c}{Sentiment (Accuracy)}\\
Model Train. Data (Num. Reviews) & \textbf{SemEval-2017} \\
\midrule
\textsc{Random} & 50.\\
\midrule
\textsc{Orig.} (3.4k) & 77.8 \\
\textsc{Orig.+Rev.} (1.7k+1.7k) & 64.2 \\
\textsc{Orig.\textsubscript{DISJOINT}+Rev.} (1.7k+1.7k) & 75.1 \\
\midrule
\textsc{Orig.} (19k) & 72.0 \\
\textsc{Orig.+Rev.} (19k+1.7k) & 66.9 \\
\textsc{Orig.\textsubscript{DISJOINT}+Rev.} (19k-1.7k+1.7k) & 76.5\\
\midrule
\textsc{Orig.} (3.4k) \textsubscript{BASE\textsubscript{uncased}} &  75.7 \\ 
\textsc{Orig.+Rev.} (1.7k+1.7k) \textsubscript{BASE\textsubscript{uncased}} & 73.5 \\
\textsc{Orig.\textsubscript{DISJOINT}+Rev.} (1.7k+1.7k) \textsubscript{BASE\textsubscript{uncased}} & 75.2 \\ 
\midrule
\textsc{Orig.} (19k) \textsubscript{BASE\textsubscript{uncased}} & 68.5 \\ 
\textsc{Orig.+Rev.} (19k+1.7k) \textsubscript{BASE\textsubscript{uncased}} & 72.6 \\
\textsc{Orig.\textsubscript{DISJOINT}+Rev.} (19k-1.7k+1.7k) \textsubscript{BASE\textsubscript{uncased}} & 76.9 \\
\bottomrule
\end{tabular}
\end{table*} 

%%%%%%%%%%%%%%%%%%%%%%%%%%%%%%%%%%%%%%%%%%
%%%% APPENDIXSECTION
%%%%%%%%%%%%%%%%%%%%%%%%%%%%%%%%%%%%%%%%%%
%\clearpage
\appendixsection{Sentiment Data: Binary Prediction of Local Annotation Edits}

Tables~\ref{table-revised-reviews-selected-awkward-phrases} and \ref{table-contrast-sets-reviews-selected-awkward-phrases} illustrate how the zero-shot sequence labeling predictions from the \textsc{uniCNN+BERT} model can be used as an assistant for analyzing text datasets, uncovering subtle patterns that are not easily discoverable in large datasets.

%\clearpage

\begin{table*}[hbt!] %[b]
\caption{Selected sentences pulled from the counterfactually-augmented dev set. \underline{Underlined} words are zero-shot sequence label predictions from the \textsc{uniCNN+BERT} model for predicting annotator domain, with correct predictions further highlighted in \textcolor{blue}{\underline{blue}} and incorrect predictions (i.e., in which the token did not participate in a ground-truth token-level diff) in \textcolor{red}{\underline{red}}. For reference, we also provide the original document of the parallel source-target pair.}
\label{table-revised-reviews-selected-awkward-phrases}
\centering
%\footnotesize
\scriptsize
\begin{tabular}{P{35mm}T{95mm}}
\toprule
\multicolumn{2}{c}{\textbf{Counterfactually-Augmented Data}}  \\
\midrule
\multicolumn{2}{c}{Review-level {\color{white}\colorbox{black}{Annotator Domain}} (\textit{Not Sentiment})}  \\
\midrule
& Dev. Set Document 40/41\\
\textsc{Original} & \ttfamily{[...] It shocks me that something exceptional like Firefly lasts one season, while garbage like the Battlestar Galactica remake spawns a spin off. [...]} \\
\textsc{uniCNN+BERT (Rev.)} & \ttfamily{[...] It shocks me that something exceptional like Firefly lasts one season, while even \textcolor{blue}{\underline{better}} shows like the Battlestar Galactica remake spawns a spin off. [...]}\\
\midrule
& Dev. Set Document 254/255\\
\textsc{Original} & \ttfamily{[...] A well made movie, one which I will always remember, and watch again.} \\
\textsc{uniCNN+BERT (Rev.)} & \ttfamily{[...] A feeble movie, one which I will always remember \textcolor{blue}{\underline{poorly,}} and never watch again.}\\
\midrule
& Dev. Set Document 258/259\\
\textsc{Original} & \ttfamily{[...] We need that time again, now more than ever. [...]} \\
\textsc{uniCNN+BERT (Rev.)} & \ttfamily{[...] We do need that time again, now \textcolor{blue}{\underline{less}} than ever. [...]}\\
\midrule
& Dev. Set Document 276/277\\
\textsc{Original} & \ttfamily{[...] Highly, hugely recommended!} \\
\textsc{uniCNN+BERT (Rev.)} & \ttfamily{[...] Highly, hugely \textcolor{blue}{\underline{not}} recommended!}\\
\midrule
& Dev. Set Document 278/279\\
\textsc{Original} & \ttfamily{almost every review of this movie I'd seen was pretty bad. It's not pretty bad, it's actually pretty good, though not great. [...]} \\
\textsc{uniCNN+BERT (Rev.)} & \ttfamily{almost every review of this movie I'd seen was pretty bad. And the reviews are correct, it's actually pretty horrible, though \textcolor{red}{\underline{not}} \textcolor{blue}{\underline{worst.}}  [...]}\\
\bottomrule
\end{tabular}
\end{table*}

\begin{table*}[hbt!] %[b]
\caption{Selected sentences pulled from the contrast sets dev set. \underline{Underlined} words are zero-shot sequence label predictions from the \textsc{uniCNN+BERT} model for predicting annotator domain, with correct predictions further highlighted in \textcolor{blue}{\underline{blue}} and incorrect predictions (i.e., in which the token did not participate in a ground-truth token-level diff) in \textcolor{red}{\underline{red}}. For reference, we also provide the original document of the parallel source-target pair.}
\label{table-contrast-sets-reviews-selected-awkward-phrases}
\centering
%\footnotesize
\scriptsize
\begin{tabular}{P{35mm}T{95mm}}
\toprule
\multicolumn{2}{c}{\textbf{Contrast Sets}}  \\
\midrule
\multicolumn{2}{c}{Review-level {\color{white}\colorbox{black}{Annotator Domain}} (\textit{Not Sentiment})}  \\
\midrule
& Dev. Set Document 38/39\\
\textsc{Original} & \ttfamily{[...] The content of the film was very very moving. [...]} \\
\textsc{uniCNN+BERT (Contrast)} & \ttfamily{[...] The content of the film was very very \textcolor{blue}{\underline{missing.}} [...]}\\
\midrule
& Dev. Set Document 58/59\\
\textsc{Original} & \ttfamily{[...] Anyone who has the slightest interest in Gaelic, folk history, folk music, oral culture, Scotland, British history, multi-culturalism or social justice should go and see this film.} \\
\textsc{uniCNN+BERT (Contrast)} & \ttfamily{[...] Anyone who has the slightest interest in Gaelic, folk history, folk music, oral culture, Scotland, British history, multi-culturalism or social justice should go and \textcolor{blue}{\underline{avoid}} this film.}\\
\midrule
& Dev. Set Document 146/147\\
\textsc{Original} & \ttfamily{[...] It is hard to describe the incredible subject matter the Maysles discovered but everything in it works wonderfully. [...]} \\
\textsc{uniCNN+BERT (Contrast)} & \ttfamily{[...] It is hard to describe the flawed subject matter the Maysles discovered but everything in it \textcolor{red}{\underline{works}} hopelessly.  [...]}\\
\midrule
& Dev. Set Document 164/165\\
\textsc{Original} & \ttfamily{[...] The characters are cardboard clichs of everything that has ever been in a bad Sci-Fi series. [...]} \\
\textsc{uniCNN+BERT (Contrast)} & \ttfamily{[...] The characters are imaginations \textcolor{red}{\underline{of}} everything that has ever been in a good Sci-Fi series. [...]}\\
\midrule
& Dev. Set Document 176/177\\
\textsc{Original} & \ttfamily{[...] There was also a forgettable sequel several years later, but this instant classic is not to be missed.} \\
\textsc{uniCNN+BERT (Contrast)} & \ttfamily{[...] There was also a \textcolor{red}{\underline{forgettable}} sequel several years later, which made this \textcolor{red}{\underline{instant}} film even more missable.}\\
\midrule
& Dev. Set Document 182/183\\
\textsc{Original} & \ttfamily{[...] It has very little plot,mostly partying,beer drinking and fighting.  [...]} \\
\textsc{uniCNN+BERT (Contrast)} & \ttfamily{[...] It has very \textcolor{blue}{\underline{dense}} plot,mostly partying,beer drinking and fighting. [...]}\\
\midrule
& Dev. Set Document 184/185\\
\textsc{Original} & \ttfamily{[...] Whatever originality exists in this film -- unusual domestic setting for a musical, lots of fantasy, some animation -- is more than offset by a script that has not an ounce of wit or thought-provoking plot development.  [...]} \\
\textsc{uniCNN+BERT (Contrast)} & \ttfamily{[...] Whatever originality exists in this film -- unusual domestic setting for a musical, lots of fantasy, some animation -- is more than offset by a script that has \textcolor{blue}{\underline{so much}} wit \textcolor{red}{\underline{or}} thought-provoking plot development. [...]}\\
\bottomrule
\end{tabular}
\end{table*}

\starttwocolumn
\bibliography{binary}

\end{document}